\documentclass[lettersize,journal]{IEEEtran}
\usepackage{amsmath,amsfonts}
\usepackage{algorithmic}
\usepackage{algorithm}
\usepackage{array}
\usepackage[caption=false,font=normalsize,labelfont=sf,textfont=sf]{subfig}
\usepackage{textcomp}
\usepackage{stfloats}
\usepackage{url}
\usepackage{verbatim}
\usepackage{graphicx}
\usepackage{cite}
\usepackage[colorlinks,urlcolor=blue,linkcolor=blue,citecolor=blue]{hyperref}
\usepackage{graphicx}

\usepackage{amssymb}
\usepackage{multirow}
\usepackage[T1]{fontenc}
\usepackage{booktabs} % for \toprule, \midrule, \bottomrule

\begin{document}

\title{Learning Interpretable Differentiable Logic Networks}
\author {
    Chang Yue and Niraj K. Jha \IEEEmembership{Fellow, IEEE}

\thanks{Chang Yue and Niraj K. Jha are with the Department of Electrical and Computer Engineering, Princeton University, 
Princeton, NJ 08544, USA, e-mail: \{cyue, jha\}@princeton.edu.}
% \thanks{This work was supported by NSF under Grant No. CNS-1907831.}
}


% % column for its text to clear the IEEEpubid mark.

\maketitle

\begin{abstract}
The ubiquity of neural networks (NNs) in real-world applications, from healthcare to 
natural language processing, underscores their immense utility in capturing complex relationships 
within high-dimensional data. However, NNs come with notable disadvantages, such as their "black-box" 
nature, which hampers interpretability, as well as their tendency to overfit the training data. 
We introduce a novel method for learning interpretable differentiable logic networks (DLNs)
that are architectures that employ multiple layers of binary logic operators. We train these networks by 
softening and differentiating their discrete components, e.g., through binarization of inputs, binary 
logic operations, and connections between neurons. This approach enables the use of gradient-based 
learning methods. Experimental results on twenty classification tasks indicate that differentiable logic 
networks can achieve accuracies comparable to or exceeding that of traditional NNs. Equally importantly, 
these networks offer the advantage of interpretability. Moreover, their relatively simple 
structure results in the number of logic gate-level operations during inference being up to 
a thousand times smaller than NNs, making them suitable for deployment on edge devices.

\end{abstract}


\section{Introduction}
Neural networks (NNs) have become a cornerstone of modern machine learning, driving 
advancements in a multitude of fields. Their impressive performance has catalyzed 
transformative changes across various sectors including healthcare and finance. Despite 
their prowess, a key drawback lies in their incomprehensible decision-making 
process often termed the "black-box" problem. This lack of interpretability 
becomes particularly salient in applications such as medical diagnostics, where 
understanding the reasoning behind the NN decisions is very important. We take inspiration 
from work on Deep Differentiable Logic Gate Networks (DDLGNs)~\cite{petersen2022deep}, 
in which the authors train classification networks consisting of logic operators 
by relaxing these operations to become differentiable. They randomly initialize a 
network such that each logic neuron is connected only to two inputs from its 
previous layer and then employ real-valued logic and relaxed logic operations 
to find the role each logic neuron should play. They can extract logic rules 
from the resulting networks, thus making the decision process 
interpretable. In some classification tasks, their model achieves comparable 
inference accuracies to regular Multi-Layer Perceptrons (MLPs) while operating 
at speeds orders of magnitude faster. In a DDLGN, the logic gate search part 
is differentiable; however, the network needs a fixed structure initialized at 
the beginning and only accepts binary inputs. We develop methods that overcome all 
these limitations by making all parts of the network differentiable. Moreover, we 
observe further performance improvement and model size reduction.

\begin{figure*}[!t]
\centering
\includegraphics[width=0.7\textwidth]{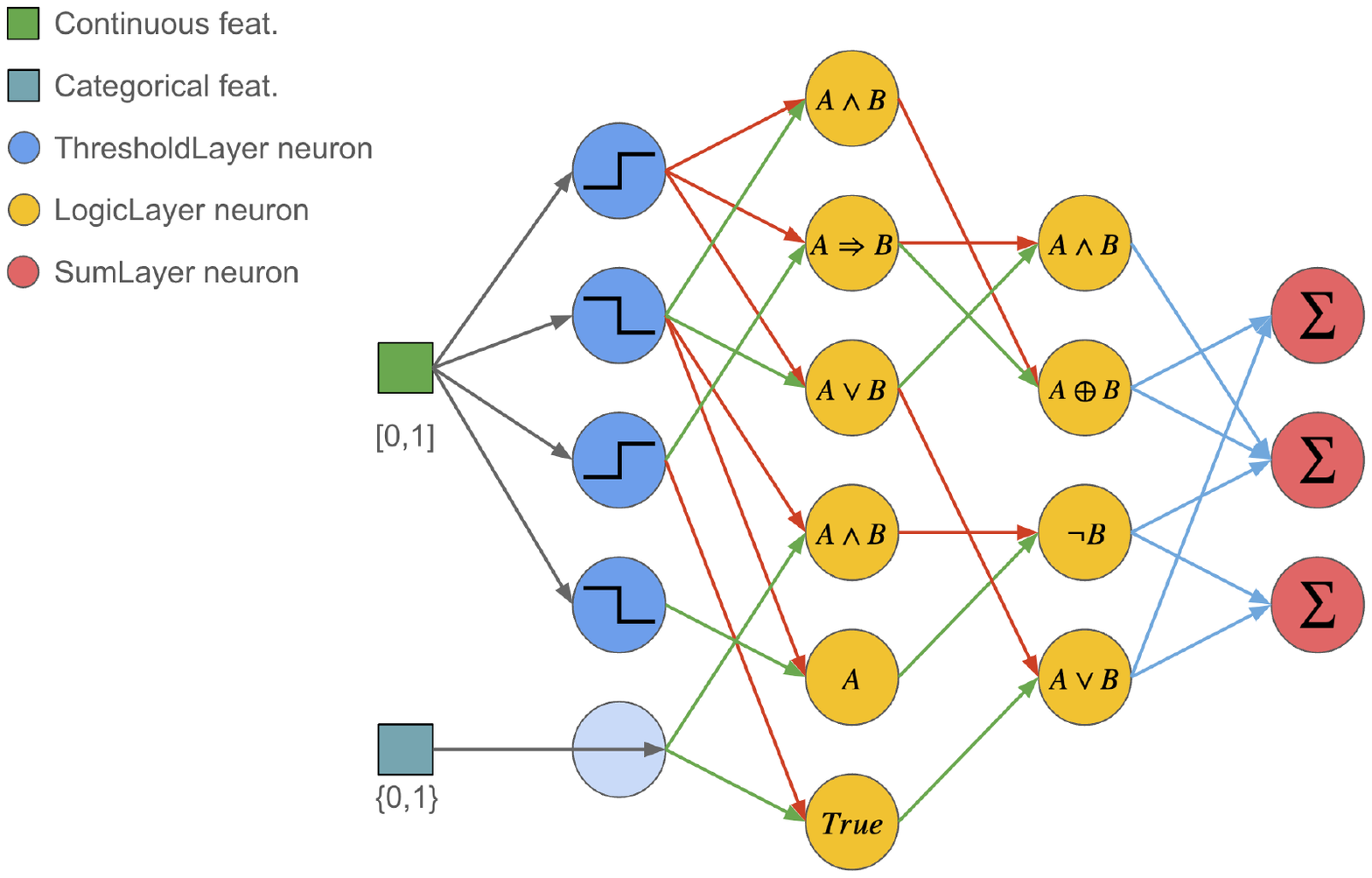}
\vspace*{-5mm}
\caption{A simplified DLN example. It takes input samples and binarizes continuous 
variables through a ThresholdLayer. It then passes the binary vector to layers of two-input Boolean 
logic operators. Finally, it counts logic rule triggers to determine the sample's class.}
\label{example}
\end{figure*}

We propose a method that can learn networks consisting of logic operators, 
trainable through gradient-based optimization. We present a simplified example 
of such a network in Fig.~\ref{example} and refer to it as a differentiable logic network (DLN) because it 
makes decisions based on logic rules. We give examples of post-processed DLNs later.
%in Figures~\ref{viz-cirrhosis},~\ref{viz-heartFailure}, and~\ref{viz-liver}. 
A DLN mainly consists of neurons that perform 
binary logic operations. Given a sample input, it first binarizes 
continuous features, then passes the binarized sample (which includes one-hot encoded
versions of categorical features) through layers of logic 
operators, and finally counts the logic rules that have been triggered to 
determine which class this sample belongs to. Our training method can find both 
the network structure, i.e., connections between layers, and the Boolean function that 
each neuron should perform. In summary, we can train interpretable logic rule-based networks 
from scratch.

Another advantage that DLNs have is computational efficiency. 
The models that perform well generally have a small number of neurons. 
Moreover, connections in DLNs are sparse.
In contrast, in traditional MLPs, neurons between layers are fully 
connected, resulting in links that are quadratic in hidden layer sizes. 
In addition, operations in our DLNs are also simple: for all neurons, 
output values are one-bit; ThresholdLayers merely perform comparisons, which 
can be implemented by a fast comparator; LogicLayer neurons only perform
basic logic operations; and SumLayer neurons only perform binary aggregation, which 
can also be efficiently implemented.

In this article, we propose a novel method for training DLNs.  The resulting network is 
accurate, interpretable, and efficient. The main contributions are as follows:
\begin{itemize}
\item{We develop methodologies to make all components of DLNs differentiable, 
making them trainable from scratch.}
\item{We present methodologies to train interpretable DLNs efficiently and to simplify them after training.}
\item{We conduct experiments and compare our model with various machine learning 
models in terms of inference accuracy and computation cost.}
\end{itemize}

The article is structured as follows. Section \ref{sec-related} discusses related work.
Section \ref{sec-method} presents the methodology. Section \ref{sec-experiments} showcases 
experimental results. Section \ref{sec-conclusion} provides a conclusion and thoughts about 
future research directions.

\section{Related Work}\label{sec-related}
\begin{table}[t]
    \centering
    \caption{List of all real-valued binary logic operations. Adapted from DDLGN~\cite{petersen2022deep}.}  
    \label{operator}
    \label{tab:operators}
    \begin{tabular}{rclccccccccc}
        \toprule
        ID    & Operator                    & Real-valued   & 00 & 01 & 10 & 11 \\
        \midrule
           0  & False                       & $0$                       & 0     & 0     & 0     & 0     \\
           1  & $A\land B$                  & $A\cdot B$                & 0     & 0     & 0     & 1     \\
           2  & $\neg(A \Rightarrow B)$     & $A-AB$                    & 0     & 0     & 1     & 0     \\
           3  & $A$                         & $A$                       & 0     & 0     & 1     & 1     \\
           4  & $\neg(A \Leftarrow B)$      & $B-AB$                    & 0     & 1     & 0     & 0     \\
           5  & $B$                         & $B$                       & 0     & 1     & 0     & 1     \\
           6  & $A \oplus B$                & $A + B - 2AB$             & 0     & 1     & 1     & 0     \\
           7  & $A \lor B$                  & $A + B - AB$              & 0     & 1     & 1     & 1     \\
           8  & $\neg(A \lor B)$            & $1 - (A + B - AB)$        & 1     & 0     & 0     & 0     \\
           9  & $\neg(A \oplus B)$          & $1 - (A + B - 2AB)$       & 1     & 0     & 0     & 1     \\
           10 & $\neg B$                    & $1 - B$                   & 1     & 0     & 1     & 0     \\
           11 & $A \Leftarrow B$            & $1-B+AB$                  & 1     & 0     & 1     & 1     \\
           12 & $\neg A$                    & $1-A$                     & 1     & 1     & 0     & 0     \\
           13 & $A \Rightarrow B$           & $1-A+AB$                  & 1     & 1     & 0     & 1     \\
           14 & $\neg(A \land B)$           & $1 - AB$                  & 1     & 1     & 1     & 0     \\
           15 & True                        & $1$                       & 1     & 1     & 1     & 1     \\
        \bottomrule
    \end{tabular}
\end{table}
In this section, we introduce techniques related to our methodologies and prior works on 
logics, differentiability, interpretable machine learning, and efficient machine learning. \\
\textbf{Logic Representation:} Boolean logic rules perform discrete operations like 
AND and XOR on binary inputs to produce binary outputs. They can be described by 
truth tables. In our designs, we limit the number of inputs to two. To remove reliance on just the $0/1$ 
values that are inherent to Boolean logic, fuzzy logic~\cite{zadeh1978fuzzy} 
can be employed to enable the truth values of variables to be any real number between 
$0$ and $1$. This interpretation~\cite{menger2003statistical, goertzel2008probabilistic} 
of logic operations extends traditional binary logic into the probabilistic realm, 
enabling differentiability. For example, $P(A \land B)$ can be realized as $P(A) \times P(B)$. We employ 
the same set of real-valued logic operations that DDLGN~\cite{petersen2022deep} uses; 
thus, we have directly adapted Table~\ref{operator} from DDLGN. 
It lists all the $2^{2^2} = 16$ Boolean functions that are possible with two inputs. \\
\textbf{Differentiability:} This is a cornerstone of gradient-based optimizations.
As discussed earlier, real-valued logics~\cite{zadeh1978fuzzy} extend binary logic to enable continuity.
Softmax is a differentiable approximation to $\text{argmax}$; it converts logits into probabilities.
Gumbel-Softmax~\cite{jang2016categorical} adds Gumbel noise to the logits before applying the Softmax
function to enable differentiable sampling. We employ the Softmax function to determine incoming links
and types of logic operators. The Straight-Through Estimator (STE)~\cite{bengio2013estimating} passes the
values of non-differentiable functions while backpropagating their approximated differentiable gradients.
It provides an alternative way to estimate gradients; we have found it to be helpful in training DLNs. \\
\textbf{Interpretable Machine Learning:} Numerous attempts have been made to explain 
the working mechanisms of black-box models, such as LIME~\cite{ribeiro2016should}, 
saliency maps~\cite{simonyan2013deep}, Shapley additive explanations~\cite{lundberg2017unified}, 
and class activation maps~\cite{zhou2016learning}. 
However, Rudin~\cite{rudin2019stop} demonstrates that these post-hoc approaches are 
often unreliable and can be misleading. A growing body of work is focused on developing 
inherently interpretable models, such as sparse decision trees~\cite{lin2020generalized} 
and scoring systems~\cite{ustun2016supersparse}. Neuro-symbolic methods bridge symbolic 
and connectionist approaches. Richardson {\em et al.}~\cite{riegel2020logical} propose 
Markov logic networks (MLNs) that combine first-order logic with probabilistic 
graphical models. MLNs employ the Markov chain Monte Carlo method for inference. However,
this approach is not scalable to large networks. Riegel {\em et al.}~\cite{richardson2006markov} 
introduce Logical Neural Networks (LNNs) that build a differentiable symbolic logic network 
first, and then learn the weights for each node. LNNs require hand-crafted symbolic graphs, 
which may necessitate extensive domain knowledge. Research efforts are also geared at learning logic 
rules using differentiable networks. Wang {\em et al.}~\cite{wang2021scalable} present
the Rule-based Representation Learner by projecting rules into a continuous space 
and propose Gradient Grafting to differentiate the network. Our methodology, on the other hand, makes all parts 
of the NN differentiable and can be optimized by vanilla gradient descent. 
Payani {\em et al.}~\cite{payani2019learning} introduce Neural Logic Networks (NLNs) to 
learn Boolean functions and develop several types of layers that perform logic operations. 
Our methodology features flexible neurons and connections. Shi \emph{et al.}~\cite{shi2019neural} 
also proposed a method named NLN that learns logic variables as vector 
representations and logic operations as neural modules, regularized by logical rules. Our 
approach learns the network structure and logic functions through relaxed differentiation, 
without any predefined modules. Another work related to our method is 
the Gumbel-Max Equation Learner~\cite{chen2020learning}. The authors apply 
continuous relaxation to the network structure via Gumbel-Max; our methodology differs in terms 
of search algorithms. In the Wide \& Deep Model~\cite{cheng2016wide} work, the authors find that 
sometimes it is better to bring the input closer to the output layer; this can also make 
the decision process more interpretable. We have found this strategy to be helpful. \\
\textbf{Efficient Machine Learning:} Techniques like pruning~\cite{han2015learning, hoefler2021sparsity}, 
quantization~\cite{courbariaux2016binarized, hubara2017quantized, gholami2022survey}, and 
knowledge distillation~\cite{hinton2015distilling, gou2021knowledge} aim to create smaller 
models that retain the performance characteristics of larger ones, although these approaches 
do not necessarily enhance model interpretability. Application-Specific Integrated Circuits 
and Field-Programmable Gate Arrays (FPGAs) sometimes offer highly specialized and 
task-specific optimizations. Umuroglu {\em et al.}~\cite{umuroglu2017finn} demonstrate that 
task-specific FPGAs can accelerate inference for binarized NNs. Our logic 
operation-based models can similarly achieve high efficiency on specialized hardware.

\section{Methodology}\label{sec-method}
In this section, we describe our method in detail, beginning with an overview, followed by 
details of each training step.

\begin{figure}[!t]
\centering
\includegraphics[width=0.3\columnwidth]{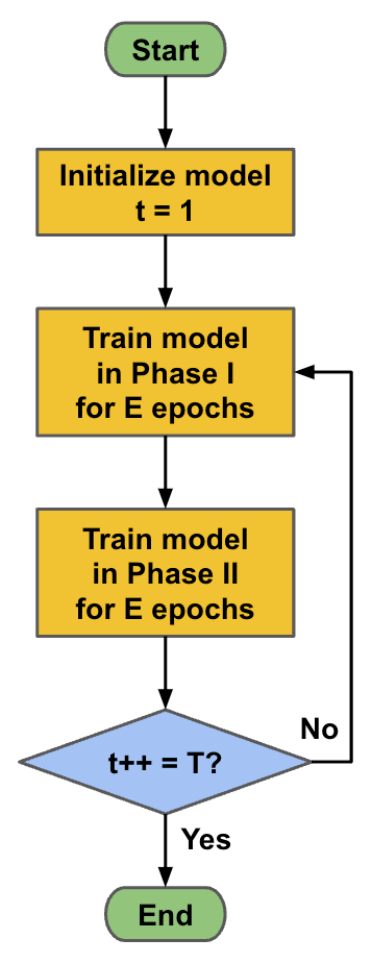}
\caption{DLN training flowchart.}
\label{flowchart}
\end{figure}

\begin{algorithm}[tb]
\caption{Skeleton training process for DLNs}\label{algo}
\begin{algorithmic}[1]
\REQUIRE{Number of iterations \( T \), epochs per phase \( E \)}
\FOR{\( t = 1 \) to \( T \)}
    \STATE{}
    \COMMENT {Phase I: optimizing neuron functions}
    \STATE{Quantize and fix neuron connections.}
    \FOR{\( \text{epoch} = 1 \) to \( E \)}
        \STATE{Apply gradient to logic gate distributions for neurons in LogicLayer and to biases and slopes for neurons in ThresholdLayer.}
        \COMMENT {Update neuron parameters}
    \ENDFOR
    \STATE{}
    \COMMENT {Phase II: optimizing neuron connections}
    \STATE{Quantize and fix functions inside neurons.}
    \FOR{\( \text{epoch} = 1 \) to \( E \)}
        \STATE{Apply gradient to find optimal connections between neurons.}
        \COMMENT {Update connection parameters}
    \ENDFOR
\ENDFOR
\end{algorithmic}
\end{algorithm}

Fig.~\ref{example} illustrates a sample DLN. It consists of three types of layers: a 
ThresholdLayer at the input to binarize continuous input features into binary $0/1$ values 
using threshold functions, multiple LogicLayers consisting of two-input logic operators for 
performing logic operations, and a SumLayer at the end to aggregate the output from the last 
LogicLayer to determine the class. We can directly extract logic rules from the network, thereby 
making its predictions interpretable.  Our DLN is similar to 
conventional MLPs in terms of input and output: both accept samples with continuous and one-hot 
categorical features and output probability predictions for each class. However, a DLN
differs from MLPs in three major ways: it transforms input into binary values at the 
first step; instead of conducting matrix multiplications between MLP hidden layers, it performs 
logic operations within the network; and the connections in LogicLayers are very sparse, 
primarily because each neuron accepts only two inputs.

Our training objectives are twofold: determining which function each neuron should implement and 
establishing how the neurons should be connected. Neuron function optimization and neuron connection 
optimization can be divided into two separate tasks because they optimize different sets of 
parameters. Therefore, we divide the training process into two phases: one for optimizing 
the internals of neurons (referred to as Phase I) and another for optimizing the connections 
between neurons (referred to as Phase II). However, both problems are discrete, making common 
NN training optimizers like gradient descent unsuitable for the task. We have 
developed methodologies that relax the discrete search spaces to make them continuous and differentiable. 
We provide details of these methodologies in subsequent subsections; a coarse summary is that we 
relax discrete search among candidates into a probabilistic mixture of candidates.

We present a high-level overview of the DLN training workflow in Fig.~\ref{flowchart} and its 
algorithmic summary in Algorithm~\ref{algo}. We iteratively update model parameters in the two phases 
in an alternating manner. After initialization, we conduct Phase I training for \(E\) epochs, 
followed by Phase II training for \(E\) epochs. We repeat the combination of Phase I and Phase II 
training for \(T\) iterations. As Algorithm~\ref{algo} suggests, during one phase, we freeze 
the trainable parameters of the other phase and perform a hard assignment to its related 
functions. Thus, in Phase I, we fix the neuron connections and take argmax over connection 
parameters; specifically, the link between any two neurons is either \(0\) or \(1\) and nothing 
in between. In Phase II, we freeze and hard-assign the neuron functions so that each neuron 
performs only one discrete logic operation among all possible operations. After freezing and quantizing 
the parameters related to the other phase, we relax the discrete parts related to the under-training 
phase to make optimization differentiable. The algorithm updates relevant parameters solely 
through gradient descent. Once training is complete, we discretize both the functions and connections 
of neurons to hard binary states, as Fig.~\ref{example} demonstrates. Generally, this discretization 
leads to a quantization loss; however, this loss is typically small compared to the generalization loss 
because most components converge well. In some cases, discretization even reduces overfitting and 
helps with generalization.

\newcommand{\cellcontent}[1]{%
  \begin{minipage}{\linewidth}
    \raggedright
    #1
  \end{minipage}}
\newcommand{\phspace}{\hspace{1em}}

\begin{table*}[t]
    \centering
    \caption{Summary of DLN trainable parameters and feedforward functions during training and 
    inference. We use \( \mathbf{x} \) to denote input and \( \mathbf{y} \) to denote the output of each layer.}
    \label{layer-params}
    \small{
    \begin{tabular}{>{\centering\arraybackslash}p{1.5cm} >{\raggedright\arraybackslash}p{4cm} >{\raggedright\arraybackslash}p{5.5cm} >{\raggedright\arraybackslash}p{5cm}}
        \toprule
          & ThresholdLayer & LogicLayer & SumLayer \\
        \midrule
        \cellcontent{Trainable\\parameters} & 
        \cellcontent{Phase I:\\
        \phspace bias \( \mathbf{b} \in \mathbb{R}^{\text{in\_dim}} \) \\
        \phspace slope \( \mathbf{s} \in \mathbb{R}^{\text{in\_dim}} \)\\
        Phase II:\\
        \phspace None} & 
        \cellcontent{Phase I:\\
        \phspace logic fn weight \( \mathbf{W} \in \mathbb{R}^{\text{out\_dim} \times 16} \) \\
        Phase II:\\
        \phspace link \(a\) weight \( \mathbf{U} \in \mathbb{R}^{\text{out\_dim} \times \text{in\_dim}} \) \\
        \phspace link \(b\) weight \( \mathbf{V} \in \mathbb{R}^{\text{out\_dim} \times \text{in\_dim}} \)} & 
        \cellcontent{Phase I:\\
        \phspace None\\
        Phase II:\\
        \phspace link weight \( \mathbf{S} \in \mathbb{R}^{\text{in\_dim} \times |\mathcal{C}|} \)} \\
        \midrule
        \cellcontent{Training\\Phase I} & 
        \cellcontent{\( \mathbf{y}_i = \text{Sigmoid}(\mathbf{s}_i \cdot (\mathbf{x}_i - \mathbf{b}_i)) \)} & 
        \cellcontent{\( \mathbf{y}_i = \sum_{k=0}^{15} \mathbf{P}_k \cdot \text{SoftLogic}_k (\mathbf{x}_{a_i}, \mathbf{x}_{b_i}) \)\\
        \( \mathbf{P} = \text{Softmax}(\mathbf{W}_{i,:}) \)\\
        \( a_i = \text{arg max}_{j} \mathbf{U}_{i,j} \)\\
        \( b_i = \text{arg max}_{j} \mathbf{V}_{i,j} \)} & 
        \cellcontent{\( \mathbf{y}_i = \sum_{j=0}^{\text{in\_dim}-1} 1_{\{ \text{Sigmoid}(\mathbf{S}_{j,i}) \ge {\theta}_{\text{sum-th}} \}} \cdot \mathbf{x}_j \)} \\
        \midrule
        \cellcontent{Training\\Phase II} &
        \cellcontent{\( \mathbf{y}_i = \text{Heaviside}(\mathbf{s}_i \cdot (\mathbf{x}_i - \mathbf{b}_i)) \)} & 
        \cellcontent{\( \mathbf{y}_i = \text{SoftLogic}_k (a, b) \)\\
        \( k = \text{arg max}_{j} \mathbf{W}_{i,j} \)\\
        \( a = \sum_{j=0}^{\text{in\_dim}-1} \left[ \text{Softmax}(\mathbf{U}_{i,:}) \right]_j \cdot \mathbf{x}_j \)\\
        \( b = \sum_{j=0}^{\text{in\_dim}-1} \left[ \text{Softmax}(\mathbf{V}_{i,:}) \right]_j \cdot \mathbf{x}_j \)} & 
        \cellcontent{\( \mathbf{y}_i = \sum_{j=0}^{\text{in\_dim}-1} \text{Sigmoid}(\mathbf{S}_{j,i}) \cdot \mathbf{x}_j \)} \\
        \midrule
        \cellcontent{Inference} & 
        \cellcontent{Same as Phase II} & 
        \cellcontent{\( \mathbf{y}_i = \text{HardLogic}_k (\mathbf{x}_{a_i}, \mathbf{x}_{b_i}) \)\\
        \( k = \text{arg max}_{j} \mathbf{W}_{i,j} \)\\
        \( a_i = \text{arg max}_{j} \mathbf{U}_{i,j} \)\\
        \( b_i = \text{arg max}_{j} \mathbf{V}_{i,j} \)} & 
        \cellcontent{Same as Phase I} \\
        \bottomrule
    \end{tabular}
    }
\end{table*}

Table~\ref{layer-params} summarizes the trainable parameters for each layer type and their usages 
in the two training phases and inference. We present the forward computation equation for each layer 
and, for simplicity, the output function at the neuron level, i.e., the function to 
calculate the \(i^{\text{th}}\) output value \( \mathbf{y}_i \) from the input vector \( \mathbf{x} \). 
The parameters involved in training are separated by phases and are applied differently between Phase I, 
Phase II, and inference. For the ThresholdLayer, we use the Heaviside function when its parameters are 
not under training and use the scaled, shifted Sigmoid function when they are under training. 
In LogicLayer, we train weights to search for optimal neuron functions and connections. 
Each weight acts as a learnable probability logit during its training phase to make the 
network differentiable. During the other phase or inference, the parameter is quantized to 
its corresponding maximum probability. In SumLayer, weights represent the connection strengths 
between the neurons of the last hidden layer and the output neurons. They are quantized to binary 
values during inference.

We present the Phase I training process in Algorithm \ref{algo-phase1}, the Phase II training 
process in Algorithm \ref{algo-phase2}, and the inference process in Algorithm \ref{algo-inf}. 
The processes for the two training phases are quite symmetric because they employ the same procedure; 
it is just that the layers switch their functions between the two modes. For clear illustration, 
we employ a mini-batch size of \(1\) and showcase the layer output computation at the neuron level, 
i.e., we show how the \(i^{\text{th}}\) output value is computed. For practical implementation, 
we use a larger batch size during training and compute the layer output with the help of vectorization. 
In the two training algorithms, we emphasize the forward pass to highlight details of how to 
make the forward propagation differentiable. Gradient-based backpropagation is conducted 
automatically as long as the forward steps are differentiable. During the forward pass, the input 
sample goes through the DLN, layer by layer; each layer takes the output of the previous 
layer as its input. Depending on the type of the layer and the phase it is in, each neuron computes 
its output based on the equations in the pseudo-code, which we discuss in more detail in the following subsections.

\subsection{Logic Operations}
Neurons in LogicLayer act as binary logic operators, which are inherently non-differentiable. 
To overcome this obstacle, we employ real-valued operations during the training phase, as outlined 
in Table~\ref{operator}. This approach effectively softens the discrete logic operations, making 
them both continuous and differentiable. For inference, we revert to the standard binary $0/1$ logic 
operations. We use \(\text{SoftLogic}_k\) to denote the real-valued version of the 
\(k^{\text{th}}\) logic operator in Table~\ref{operator} and \(\text{HardLogic}_k\) to denote its 
binary operation. Consider the example of an AND logic gate, which is described first in the 
table. For this gate, the soft (real-valued) and hard (binary) outputs are obtained as follows. For the 
soft output, the real-valued function is:
\[
\begin{aligned}
\text{real-valued AND} (a, b) &= \text{SoftLogic}_1 (a, b) \\
&= a \cdot b,
\end{aligned}
\]
where \( a, b \in [0, 1] \) are continuous real numbers.
For the hard output, the discretized function is:
\[
\begin{aligned}
\text{binary AND} (a, b) &= \text{HardLogic}_1 (a, b) \\
&= 1_{\{ a = 1\}} \cdot 1_{\{ b = 1\}},
\end{aligned}
\]
where \( a, b \in \{0, 1\} \) are discrete binary numbers, and indicator functions are applied to them.

To ensure gradient flow between layers, we utilize fuzzy logic expressions throughout the training 
process, regardless of the phase type. As shown in line~\ref{algo-phase1-softlogic} of 
Algorithm~\ref{algo-phase1} and line~\ref{algo-phase2-softlogic} of Algorithm~\ref{algo-phase2}, 
we use \(\text{SoftLogic}\) in both training Phase I and Phase II. Line~\ref{algo-inf-hardlogic} 
of Algorithm~\ref{algo-inf} indicates that we use \(\text{HardLogic}\) for inference.

\begin{algorithm}[tb]
\caption{Phase~I training}\label{algo-phase1}
\begin{algorithmic}[1]
    \REQUIRE Training dataset \( \mathcal{D} \), SumLayer link threshold \( {\theta}_{\text{sum-th}} \)
    \ENSURE Trained model parameters \( \mathbf{b}, \mathbf{s}, \mathbf{W} \)
    \STATE Freeze link parameters \( \mathbf{U}, \mathbf{V}, \mathbf{S} \)
    \FOR{\( \text{epoch} = 1, 2, \ldots, E \)}
        \FORALL{\( (\mathbf{x}, z) \) in \( \mathcal{D} \)}
            \STATE \COMMENT{Forward pass}
            \FOR{layer in model\_layers}
                \IF{\( \text{layer} = \text{ThresholdLayer} \)} \label{algo-phase1-thresh-start}
                    \STATE \COMMENT {Trainable parameters: \( \mathbf{b}, \mathbf{s} \)}
                    \STATE \( \mathbf{y}_i = \text{Sigmoid}(\mathbf{s}_i \cdot (\mathbf{x}_i - \mathbf{b}_i)) \) \label{algo-phase1-thresh-end}
                \ELSIF{\( \text{layer} = \text{LogicLayer} \)}
                    \STATE \COMMENT {Trainable parameters: \( \mathbf{W} \)}
                    \STATE \( \mathbf{P} = \text{Softmax}(\mathbf{W}_{i,:}) \) \label{algo-phase1-logic-prob}
                    \STATE \( a_i = \text{arg max}_{j} \, \mathbf{U}_{i,j} \) \label{algo-phase1-a}
                    \STATE \( b_i = \text{arg max}_{j} \, \mathbf{V}_{i,j} \) \label{algo-phase1-b}
                    \STATE \( \mathbf{y}_i = \sum_{k=0}^{15} \mathbf{P}_k \cdot \text{SoftLogic}_k (\mathbf{x}_{a_i}, \mathbf{x}_{b_i}) \) \label{algo-phase1-softlogic} \label{algo-phase1-logic-sum}
                \ELSE
                    \STATE \COMMENT {Trainable parameters: None}
                    \STATE \( \mathbf{y}_i = \sum_{j=0}^{\text{in\_dim}-1} 1_{\{ \text{Sigmoid}(\mathbf{S}_{j,i}) \ge {\theta}_{\text{sum-th}} \}} \cdot \mathbf{x}_j \) \label{algo-phase1-sum}
                \ENDIF
                \STATE \( \mathbf{x} = \mathbf{y} \) \COMMENT {Take preceding layer's output as input}
            \ENDFOR
            \STATE{} \COMMENT{Backward pass}
            \STATE Compute loss \( \mathcal{L}(\mathbf{x}, z) \)
            \STATE Backpropagate to compute gradients
            \STATE Update neuron function parameters \( \mathbf{b}, \mathbf{s}, \mathbf{W} \)
        \ENDFOR
    \ENDFOR
\end{algorithmic}
\end{algorithm}

\begin{algorithm}[tb]
\caption{Phase~II training}\label{algo-phase2}
\begin{algorithmic}[1]
    \REQUIRE Training dataset \( \mathcal{D} \)
    \ENSURE Trained model parameters \( \mathbf{U}, \mathbf{V}, \mathbf{S} \)
    \STATE Freeze neuron function parameters \( \mathbf{b}, \mathbf{s}, \mathbf{W} \)
    \FOR{\( \text{epoch} = 1, 2, \ldots, E \)}
        \FORALL{\( (\mathbf{x}, z) \) in \( \mathcal{D} \)}
            \STATE \COMMENT{Forward pass}
            \FOR{layer in model\_layers}
                \IF{\( \text{layer} = \text{ThresholdLayer} \)} \label{algo-phase2-thresh-start}
                    \STATE \COMMENT {Trainable parameters: None}
                    \STATE \( \mathbf{y}_i = \text{Heaviside}(\mathbf{s}_i \cdot (\mathbf{x}_i - \mathbf{b}_i)) \) \label{algo-phase2-thresh-end}
                \ELSIF{\( \text{layer} = \text{LogicLayer} \)}
                    \STATE \COMMENT {Trainable parameters: \( \mathbf{U}, \mathbf{V} \)}
                    \STATE \( k = \text{arg max}_{j} \, \mathbf{W}_{i,j} \) \label{algo-phase2-logic-idx}
                    \STATE \( a = \sum_{j=0}^{\text{in\_dim}-1} \left[ \text{Softmax}(\mathbf{U}_{i,:}) \right]_j \cdot \mathbf{x}_j \) \label{algo-phase2-a}
                    \STATE \( b = \sum_{j=0}^{\text{in\_dim}-1} \left[ \text{Softmax}(\mathbf{V}_{i,:}) \right]_j \cdot \mathbf{x}_j \) \label{algo-phase2-b}
                    \STATE \( \mathbf{y}_i = \text{SoftLogic}_k (a, b) \) \label{algo-phase2-softlogic} \label{algo-phase2-logic-op}
                \ELSE
                    \STATE \COMMENT {Trainable parameters: \( \mathbf{S} \)}
                    \STATE \( \mathbf{y}_i = \sum_{j=0}^{\text{in\_dim}-1} \text{Sigmoid}(\mathbf{S}_{j,i}) \cdot \mathbf{x}_j \) \label{algo-phase2-sum}
                \ENDIF
                \STATE \( \mathbf{x} = \mathbf{y} \) \COMMENT {Take preceding layer's output as input}
            \ENDFOR
            \STATE{} \COMMENT{Backward pass}
            \STATE Compute loss \( \mathcal{L}(\mathbf{x}, z) \)
            \STATE Backpropagate to compute gradients
            \STATE Update link parameters \( \mathbf{U}, \mathbf{V}, \mathbf{S} \)
        \ENDFOR
    \ENDFOR
\end{algorithmic}
\end{algorithm}

\begin{algorithm}[!htbp]
\caption{Inference}\label{algo-inf}
\begin{algorithmic}[1]
    \REQUIRE Input data \( \mathbf{x} \)
    \REQUIRE Model layer list \( \left[ \begin{aligned} &\text{ThresholdLayer} \\
                                                        & \text{LogicLayer}_1 \\
                                                        & \ldots \\
                                                        & \text{LogicLayer}_n \\
                                                        &\text{SumLayer} \end{aligned} \right] \)
    \STATE \COMMENT{First layer: ThresholdLayer}
    \STATE \( \mathbf{y}_i = \text{Heaviside}(\mathbf{s}_i \cdot (\mathbf{x}_i - \mathbf{b}_i)) \) \label{algo-inf-thresh}
    \STATE \COMMENT {Middle layers: LogicLayer}
    \FOR {\( l = 1, 2, \ldots, n \)}
        \STATE \( k = \text{arg max}_{j} \, \mathbf{W}_{i,j} \) \label{algo-inf-logic-idx}
        \STATE \( a_i = \text{arg max}_{j} \, \mathbf{U}_{i,j} \) \label{algo-inf-a}
        \STATE \( b_i = \text{arg max}_{j} \, \mathbf{V}_{i,j} \) \label{algo-inf-b}
        \STATE \( \mathbf{x} = \mathbf{y} \)  \COMMENT {Take preceding layer's output as input}
        \STATE \( \mathbf{y}_i = \text{HardLogic}_k (\mathbf{x}_{a_i}, \mathbf{x}_{b_i}) \) \label{algo-inf-hardlogic} \label{algo-inf-logic-op}
    \ENDFOR
    \STATE \COMMENT {Last layer: SumLayer}
    \STATE \( \mathbf{z}_i = \sum_{j=0}^{\text{in\_dim}-1} 1_{\{ \text{Sigmoid}(\mathbf{S}_{j,i}) \ge {\theta}_{\text{sum-th}} \}} \cdot \mathbf{y}_j \) \label{algo-inf-sum}
    \STATE \COMMENT {Get predicted label}
    \STATE \( \hat{y} = \text{arg max}_i \mathbf{z}_i \)
    \RETURN \( \hat{y} \)
\end{algorithmic}
\end{algorithm}

\begin{figure*}[!htbp]
\centering
\includegraphics[width=\textwidth]{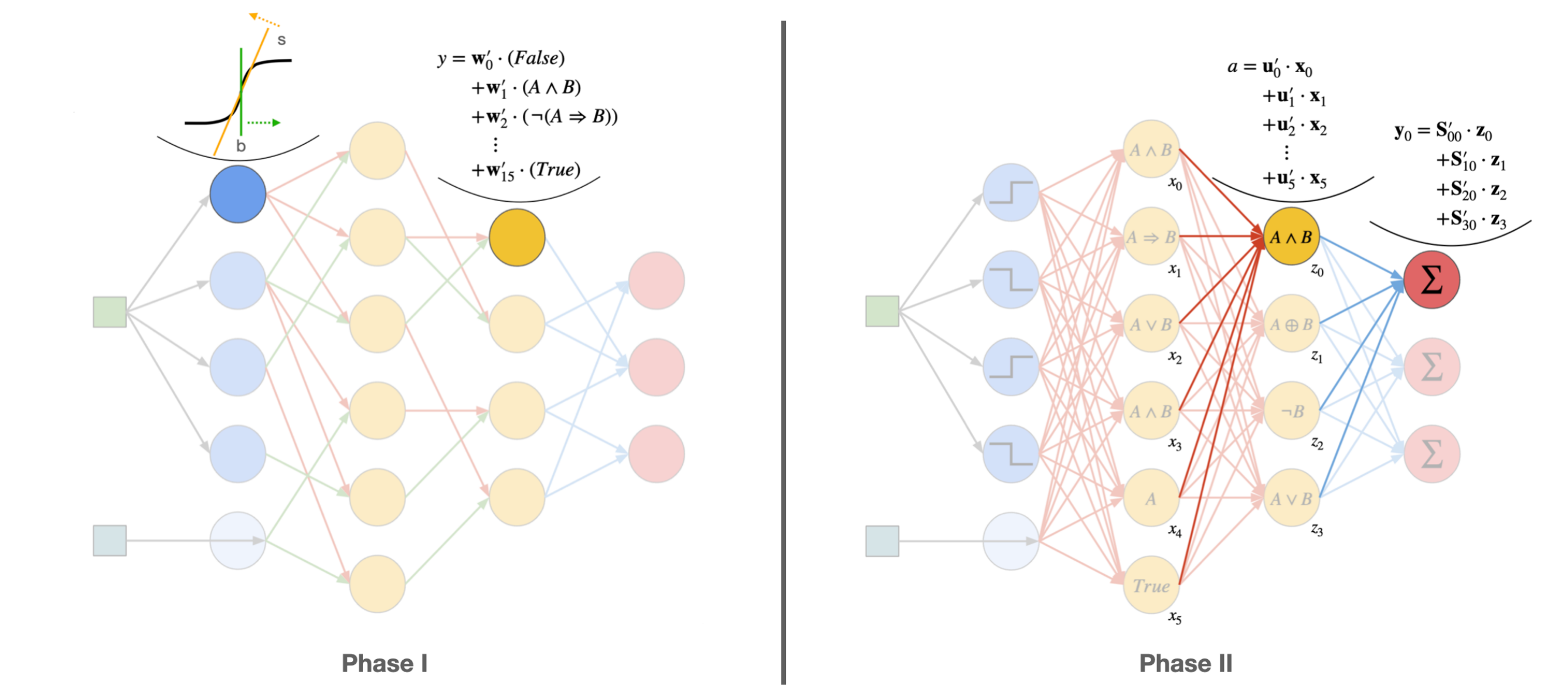}
\caption{Illustration of the two-phase training algorithm. During Phase~I, we discretize and fix 
the connections between neurons while training their parameters within ThresholdLayer and LogicLayer. 
In Phase~II, we hold the neuron operations constant and focus on optimizing the connections between them. 
Key details are highlighted for clarification.}
\label{training}
\end{figure*}

\subsection{Phase I: Determining Neuron Functions}
The left-hand side of Fig.~\ref{training} illustrates the process of determining the functions of 
neurons. In this phase, the connections between neurons are discretized and frozen, allowing us to 
focus on training the neuron parameters. DLNs consist of three types of layers: 
ThresholdLayer, LogicLayer, and SumLayer. Correspondingly, there are three types of neurons. 
Neurons in the SumLayer simply aggregate the outputs from the preceding LogicLayer that connect 
to them; therefore, they do not require any parameters to describe their function. During this phase, 
we focus on training the parameters for neurons in ThresholdLayer and LogicLayer. In the 
following subsections, we detail the training algorithms for each layer type.

\subsubsection{ThresholdLayer}
The purpose of the ThresholdLayer is to binarize continuous variables using threshold functions. 
This layer takes a continuous vector as input and outputs a binary vector. Each neuron takes one 
scalar input and outputs a scalar value. The output binary vector can later be concatenated with 
one-hot encoded categorical feature values to form the final binary vector for the next layer. 
We connect each continuous input value to multiple neurons to map it to a finite set of 
values, an approach similar to binning but not limited to mutually exclusive bins. Each neuron 
takes in one scalar and serves as a threshold function with trainable bias and scale. 
During training, we soften the non-differentiable threshold function by substituting it with 
a shifted and scaled sigmoid function:
\[
\begin{aligned}
f(x) &= \text{Sigmoid} (s \cdot (x - b)) \\
&= \frac{1}{1 + e^{- [s \cdot (x - b)]}},
\end{aligned}
\]
where \( b \) represents the cutoff and \( s \) denotes the slope of the input passed to the 
Sigmoid function. This modification enables gradient flow to both \( b \) and \( s \). 
During Phase I of training, both \( b \) and \( s \) are trainable. During Phase II or inference, 
we quantize the soft threshold function to the normal threshold function: 
\[
F(x) = \text{Heaviside} (s \cdot (x - b))
\]
and freeze parameters \( b \) and \( s \). In Fig.~\ref{training}, we illustrate the training 
mode of the ThresholdLayer on the left and the inference mode on the right (ThresholdLayer 
functions the same way in both Phase II and during inference). In Phase I, the highlighted 
neuron applies a shifted and scaled Sigmoid function to its input. After forward and 
backward propagations, gradients suggest a positive increase in \(s\) (which results in steeper 
scaling) and a positive increase in \(b\) (which results in a rightward shift of the threshold 
function). When in Phase II, the function of this neuron is quantized and fixed to a Heaviside 
function and is no longer trainable. We show the Phase I training pseudo-code for the ThresholdLayer
in lines~\ref{algo-phase1-thresh-start} to \ref{algo-phase1-thresh-end} in Algorithm~\ref{algo-phase1}, 
the Phase II training pseudo-code in lines~\ref{algo-phase2-thresh-start} to \ref{algo-phase2-thresh-end} 
in Algorithm~\ref{algo-phase2}, and the inference pseudo-code in line~\ref{algo-inf-thresh} in Algorithm~\ref{algo-inf}.

\begin{figure}[!t]
\centering
\includegraphics[width=\columnwidth]{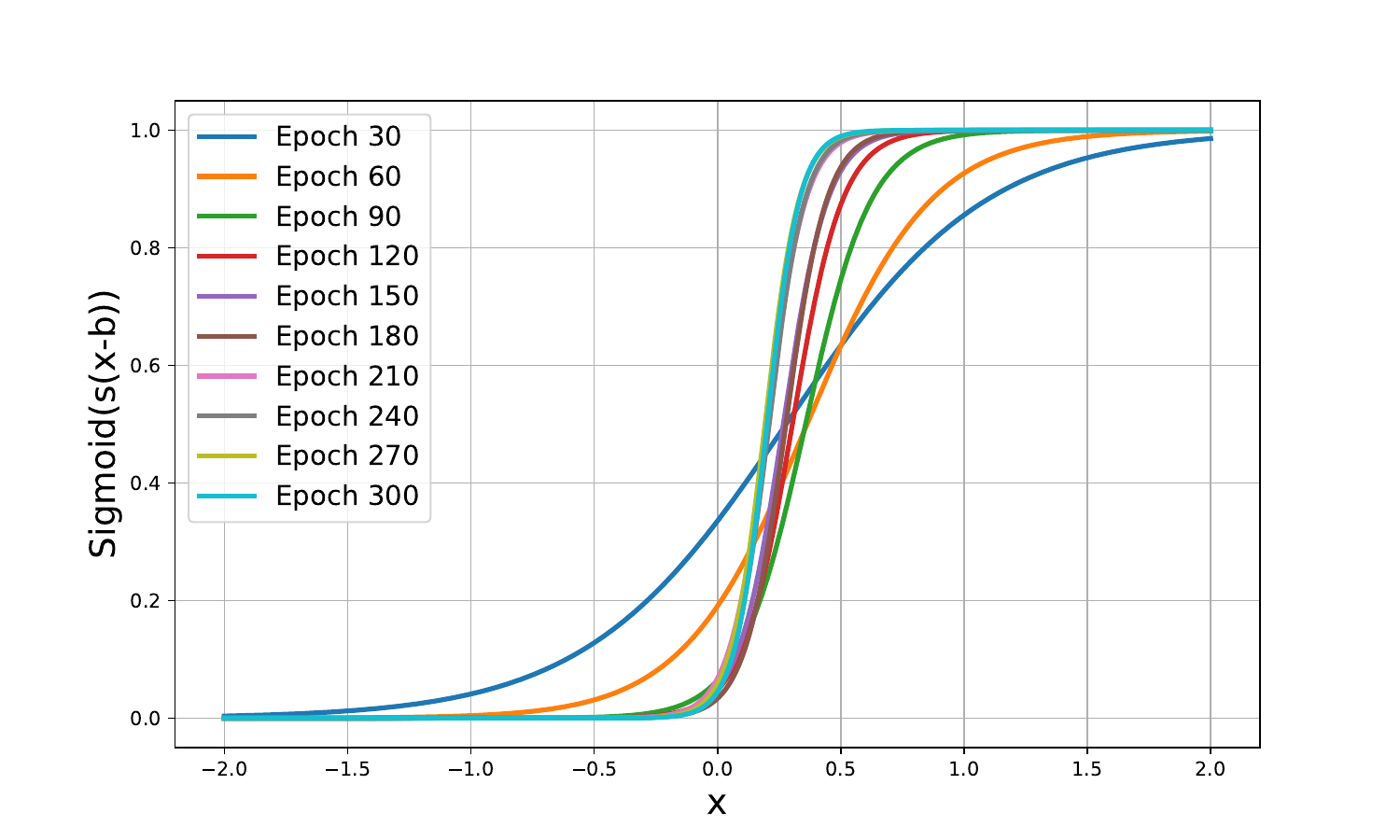}
\caption{Example training process of a ThresholdLayer neuron.}
\label{th-neuron}
\end{figure}

In our experiments, we preprocess the input data by applying Min-Max Scaling to continuous 
variables and one-hot encoding to categorical variables, ensuring that all input values 
range between \(0\) and \(1\). We allocate four or six neurons to each continuous input and 
initialize the starting slope to \(2\), a value not too flat but still smooth in the beginning. 
Inspired by Gorishniy {\em et al.}~\cite{gorishniy2022embeddings} who apply decision tree-based binning 
to transform the scalar continuous features to vectors, we initialize \(b\) for each neuron by the 
bin edges of calculated decision trees. Hence, we fit a decision tree using training data, by 
specifying the maximum number of leaf nodes. We have observed that most neurons in the ThresholdLayer 
converge during training, with some converging to values within the \(0\) and \(1\) range while others do not. 
Neurons whose biases converge to below \(0\) or over \(1\) act as constant Boolean False or True.
We found that out-of-range convergence leads to good feature selection. 
As we show in Sec.~\ref{exp-ablation}, DLNs with fixed (i.e., non-trainable) ThresholdLayers
underperform DLNs with trainable ThresholdLayers and are larger in size. Fig.~\ref{th-neuron} shows an example 
of a neuron's subsampled training process. It plots the resulting neuron function based on the 
\(b\) and \(s\) values at the sampled epochs. In the beginning, it starts with a certain cutoff and a small slope, 
thus initially acting as a smooth function. As the epochs progress, both the cutoff and slope 
converge, with the bias shifting to the left and the slope increasing, making the neuron 
increasingly resemble a threshold function.

\subsubsection{LogicLayer}
As indicated in Table~\ref{operator}, a two-input neuron can implement $2^{2^2} = 16$ Boolean functions.
Next, our goal is to determine which of the $16$ classes each neuron should belong to. This can be 
treated as a classification problem. We employ the method introduced in the DDLGN 
article~\cite{petersen2022deep}. During training, each neuron is considered a weighted sum of 
all $16$ operators. This weight is trainable. To make sure that the weights associated with the operators 
sum to $1$, a Softmax function is applied to the weights. Let a 16-valued weight vector $\mathbf{w}$ 
denote the weights for a sample logic neuron. Thus, $\mathbf{w}_i$ corresponds to the \(i^{\text{th}}\) 
Boolean operator. In this phase, connections to neurons are fixed, with each logic neuron receiving 
inputs from exactly two input neurons. Given scalar inputs $a$ and $b$, the output \(y\) of 
the logic neuron is given by the sum of logic operator mixtures:
\[
\begin{aligned}
y &= \sum_{k=0}^{15} \left[ \text{Softmax}(\mathbf{w}) \right]_k \cdot \text{SoftLogic}_k (a, b) \\
&= \sum_{k=0}^{15} \frac{e^{\mathbf{w}_k}}{\sum_j e^{\mathbf{w}_j}} \cdot \text{SoftLogic}_k (a, b).
\end{aligned}
\]
Lines \ref{algo-phase1-logic-prob} to \ref{algo-phase1-logic-sum} of Algorithm \ref{algo-phase1} 
calculate the weights and sum up the $16$ logic function outputs.

During Phase II or inference, we quantize the summation by selecting the one operator that has the 
largest corresponding weight. We find the index of this operator and employ solely its operation 
as the logic neuron's function:
\[
\begin{aligned}
k &= \arg\max_{j} \, \mathbf{w}_j, \\
y &= 
\begin{cases} 
\text{SoftLogic}_k (a, b), & \text{if training}, \\
\text{HardLogic}_k (a, b), & \text{otherwise}.
\end{cases}
\end{aligned}
\]
In Algorithm \ref{algo-phase2}, line \ref{algo-phase2-logic-idx} finds the index, and 
line \ref{algo-phase2-logic-op} performs the real-valued logic operation. In the inference 
algorithm, Algorithm \ref{algo-inf}, line \ref{algo-inf-logic-idx} finds the index, and 
line \ref{algo-inf-logic-op} performs the binary logic operation. We illustrate in 
Fig.~\ref{training} the training mode of the LogicLayer neuron function on the left and 
the inference mode on the right. Neuron function training only occurs in Phase I; hence,
neurons perform the same function in Phase II and during inference. In Phase I, the 
output of a neuron is a weighted sum of all $16$ logic functions, and this weight is 
trainable. When not under training, the function of the neuron is frozen and attached 
to one type of operator.

\begin{figure}[!t]
\centering
\includegraphics[width=\columnwidth]{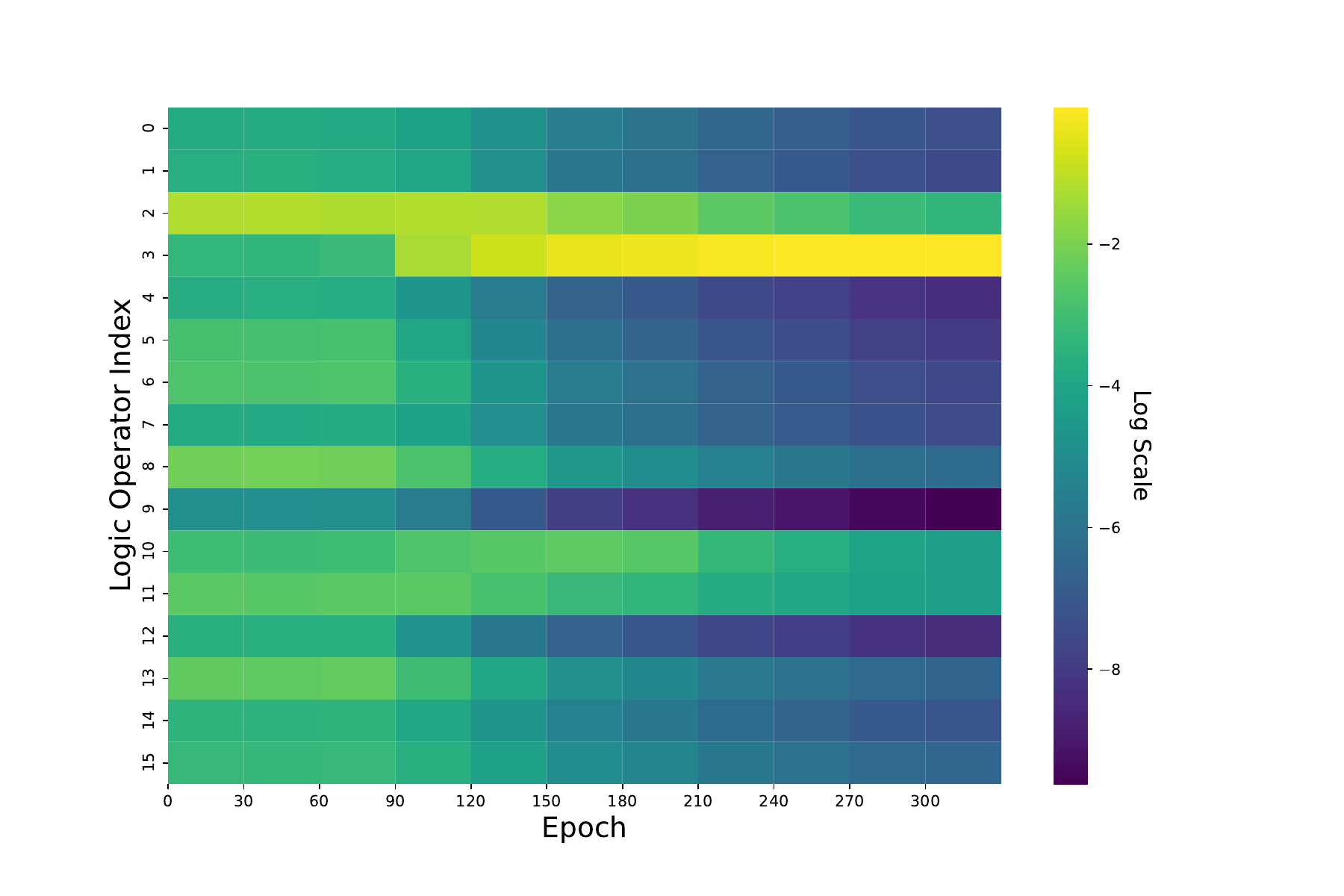}
\caption{Example training of a LogicLayer neuron. Note that colors are displayed on a logarithmic scale.}
\label{lg-neuron}
\end{figure}

In practice, with the aid of the search for neuron connections in Phase II, we find that 
most neurons can converge to a single operator. In Fig.~\ref{lg-neuron}, we show a typical 
example of a logic neuron's subsampled training process. The \(x\)-axis displays the 
epoch number and the \(y\)-axis shows the softmaxed weight distribution of the $16$ 
operators on a logarithmic color scale. As illustrated in the figure, this neuron rapidly 
converges to the fourth (index no. 3, shown in yellow) logic operator.

\subsection{Phase II: Determining Connections}
The right-hand side of Fig.~\ref{training} illustrates the training process for determining 
the connections between neurons. In this phase, we fix the functions of neurons; that is, 
the parameters of the threshold functions for threshold neurons and the selection of logic 
operators for logic neurons. For the ThresholdLayer, the input connections are fixed after
initialization. In our setup, each continuous input feature is connected to several threshold 
neurons, and each neuron takes only one input feature. Therefore, the ThresholdLayer does 
not have trainable parameters in Phase II. Next, we describe how 
connections to the LogicLayers and SumLayer are determined.

\subsubsection{LogicLayer}
Each neuron in a LogicLayer has two inputs, denoted as \(a\) and \(b\). The objective is 
to determine which two input values, among \( \text{in\_dim} \) input values from the previous 
layer's output, should be linked to \(a\) and \(b\). We divide this task into two sub-goals: 
determining the source of \(a\) and determining the source of \(b\). During training, we soften 
the connections by representing them as weighted sums of all inputs. We showcase our method 
for finding the first input, \(a\), to the logic neuron. The method for finding a connection 
to \(b\) is the same. Let \(\mathbf{x} \in \mathbb{R}^{\text{in\_dim}}\) represent the input 
of the logic neuron under study. In a LogicLayer, there are \(\text{out\_dim}\) neurons; 
suppose we pick a random neuron for study. By design, the input \(a\) of this neuron must 
be one of the elements in \(\mathbf{x}\). Hence, the goal is to find the index of that element. 
This is a discrete optimization problem. During Phase II training, we relax the problem through
weighted connections to all elements in \(\mathbf{x}\) and reduce the problem to 
training a weight vector \(\mathbf{u} \in \mathbb{R}^{\text{in\_dim}}\) to learn the weights 
linking the logic neuron to its input. We use a Softmax function to make sure that the weights 
sum to \(1\). The equation for the relaxed \(a\) is then a probabilistic mixture of \(\mathbf{x}\):
\[
\begin{aligned}
a &= \sum_{j=0}^{\text{in\_dim}-1} \left[ \text{Softmax}(\mathbf{u}) \right]_j \cdot \mathbf{x}_j \\
&= \sum_{j=0}^{\text{in\_dim}-1} \frac{e^{\mathbf{u}_j}}{\sum_i e^{\mathbf{u}_i}} \cdot \mathbf{x}_j .
\end{aligned}
\]
This approach allows each neuron to be non-uniformly connected to all neurons from the 
preceding layer during training. In practice, the neuron learns the importance of incoming 
connections and typically converges to the most relevant one.

During Phase I or inference, the connections are quantized by selecting the most heavily 
weighted input neuron, resulting in
\[
\begin{aligned}
k &= \text{arg max}_{j} \, \mathbf{u}_{j}, \\
a &= \mathbf{x}_k .
\end{aligned}
\]
The left side of Fig.~\ref{training} illustrates fixed connections to logic neurons, 
with \(a\) colored in red and \(b\) colored in green. Each neuron takes only one input as \(a\) and 
one as \(b\). Connections are discretized and fixed during both Phase I training and inference. 
Lines \ref{algo-phase1-a} and \ref{algo-phase1-b} in Algorithm \ref{algo-phase1} show 
discrete link selection during Phase I training. Lines \ref{algo-inf-a} and \ref{algo-inf-b} 
in Algorithm \ref{algo-inf} show the same argmax selections during inference. On the right side of 
Fig.~\ref{training}, we highlight an example of a connection weight matrix during training. 
In this phase, the neuron's incoming \(a\) is connected to all its inputs, and the value of 
the final \(a\) is a weighted sum of all these candidates. We expect gradient descent steps 
to make the weight concentrate on one link in the end. Lines \ref{algo-phase2-a} and \ref{algo-phase2-b} 
in Algorithm \ref{algo-phase2} show the relaxed, differentiable link search for logic neurons during
Phase II training.

\subsubsection{SumLayer}
The SumLayer has a number of neurons equal to the number of classes, \(|\mathcal{C}|\), with each 
neuron representing a class. This setup is similar to the output layer of an MLP. Each neuron 
in the SumLayer connects to a subset of neurons from the previous layer and sums up their 
output scores. These scores are normalized, then used as logits, and later passed to 
the Cross-Entropy Loss function for training. To maximize the expressiveness of the network, 
we allow nodes in the last hidden layer to connect to multiple neurons in the SumLayer. 
Consequently, this setup translates to a classification 
problem for each (input, output) neuron pair: determine whether they should connect or not. 
We train a matrix \(\mathbf{S} \in \mathbb{R}^{\text{in\_dim} \times |\mathcal{C}|}\) and treat 
\(\text{Sigmoid}(\mathbf{S}_{j,i})\) as the probability that the \(j^{\text{th}}\) input neuron 
should connect with the \(i^{\text{th}}\) output class. In this training phase, for any output 
neuron, we sum up probabilistic connections from all inputs to enable differentiability, 
as illustrated on the right side of Fig.~\ref{training}. 

Suppose we pick the \(i^{\text{th}}\) output neuron to study. Mathematically, we compute its output 
\(\mathbf{y}_i\) from input \(\mathbf{x} \in \mathbb{R}^{\text{in\_dim}}\) using:
\[
\begin{aligned}
\mathbf{y}_i &= \sum_{j=0}^{\text{in\_dim}-1} \text{Sigmoid}(\mathbf{S}_{j,i}) \cdot \mathbf{x}_j .
\end{aligned}
\]
Line~\ref{algo-phase2-sum} of Algorithm~\ref{algo-phase2} implements this equation.

When not in training mode, i.e., in Phase I or in inference, the output logit \(\mathbf{y}_i\) 
is the binary sum of inputs that have connection probability above a threshold, namely 
\({\theta}_{\text{sum-th}}\), as illustrated on the left side of Fig.~\ref{training}. 
Mathematically,
\[
\begin{aligned}
\mathbf{y}_i &= \sum_{j=0}^{\text{in\_dim}-1} 1_{\{ \text{Sigmoid}(\mathbf{S}_{j,i}) \ge {\theta}_{\text{sum-th}} \}} \cdot \mathbf{x}_j .
\end{aligned}
\]
We show the pseudo-code for this equation in line~\ref{algo-phase1-sum} of Algorithm~\ref{algo-phase1} 
and in line~\ref{algo-inf-sum} of Algorithm~\ref{algo-inf}. We impose binary connections to SumLayers 
to maximize interpretability. We found that setting \({\theta}_{\text{sum-th}}\) to \(0.8\) works well, in
general.

\subsection{Searching over Subspaces} \label{method-subspace}
In LogicLayers, for each neuron, the full search space has a size of \(16\) for the neuron type and a size 
of \(\text{in\_dim}\) for the incoming link. 
However, full search is often not necessary.
For example, any logic circuit can be constructed using just NAND gates. Selecting 
a subset of gates simplifies the training objective and helps with the vanishing gradient problem, as 
it reduces the weight dimension in Softmax layers. We could select only logic gates that exhibit good 
gradient flow. From Table \ref{operator}, we can see that the OR gate has better gradient flow than the 
AND gate because inputs \(A\) and \(B\) are between \(0\) and \(1\). A similar situation applies to link 
search: there are many paths in a network that can yield good local optima. Fortunately, searching over 
a subspace can be flexibly implemented, thanks to the properties of the Softmax function. We just 
need to assign \(-\infty\) to the weight entries that we want to turn off the search for; their Softmax 
probabilities will then be zero. Thus, they will never be active during training and inference, and 
there is no gradient flow to them.

We sort the gate types by logic completeness and gradient flow, from high to low: NOR, NAND, XOR, XNOR, OR, AND, 
A, B, NOT A, NOT B, B IMPLIES A, A IMPLIES B, NOT (A IMPLIES B), NOT (B IMPLIES A), False, True. We can just include
the first \(k\) candidates from the list and mask out the others. To obtain a subset of the link space, we randomly 
select \(k\) incoming candidates and mask out the weights of the other candidates by assigning them \(-\infty\). 
In practice, we found that using a subset of \(8\) logic operators and a subset of \(8\) links works the best.
Thus, we use this setting as the default. We present the results of not using subsets, using a subset size of \(16\) for 
link search, and using a subset size of \(4\) for both gates and links search in Sec.~\ref{exp-ablation}.

\subsection{Using Straight-Through Estimators} \label{method-ste}
By design, every layer in a DLN employs either a Sigmoid or Softmax function. This enables 
differentiability of the network. However, although Sigmoid and Softmax functions soften the step 
and argmax functions, they also weaken output activations. After encountering layers of Sigmoid and Softmax functions, 
the layer outputs tend towards the average and become more uniform, making the gradient descent method harder to 
work with. To sharpen the function outputs while enabling backpropagation of gradients at the 
same time, we employ the Straight-Through Estimator (STE). During the forward pass, STE applies 
the non-differentiable function (e.g., threshold, argmax, etc.) 
normally, but during the backward pass, it bypasses the non-differentiable 
function by treating its derivative as \(0\), allowing gradients from the subsequent layer to pass through 
unchanged. Mathematically, assume \(f\) is the discretized output value and \(g\) is the soft value; the 
final output \(h\) then is 
\[
\begin{aligned}
h &= \text{detach}(f - g) + g ,
\end{aligned}
\]
where \text{detach} is an operator that detaches the value from the backpropagation graph. Thus, 
the value \(f\) is passed to \(h\) but only \(g\) is counted for calculating gradients. This enables the 
output to attain the same value as the discrete function during the forward pass, and the same gradient 
as the soft function during the backward pass. 

In practice, we find STE useful for DLN training and have set its use as the default. We provide 
results for the case when STE is not used in Sec.~\ref{exp-ablation}.

% Placing the table here to make it appear on the same page as the start of the Experiments section
\begin{table*}[!htb]
    \centering
    \caption{Characteristics of the post-processed datasets and the average balanced-class test accuracy of models} \label{datasets-stats}\label{results-acc}
    \footnotesize{
    \begin{tabular}{c|ccccc|cccccccc}
      & \#  & \# & \# & \# & \# & \multirow{2}*{KNN} & \multirow{2}*{DT} & \multirow{2}*{RF} & \multirow{2}*{NB} & \multirow{2}*{AB} & \multirow{2}*{MLP} & \multirow{2}*{LGN} & \multirow{2}*{DLN} \\
      Dataset  & train & test & class & cont. & cate. & & & & &  & &  & \\
    \hline
     BankChurn        &        7500 &       2500 &           2 &         6 &         5 &  65.1  & 69.5  & 71.6 & 64.4 &      70.7  &  74.9  &   75.6  &  75.8  \\
     BreastCancer     &         214 &         72 &           2 &         0 &        34 &  61.4  & 62.1  & 66.1 & 57.1 &      56.6  &  60.2  &   62.4  &  65.1  \\
     Cardiotocography &        1594 &        532 &           3 &        19 &        22 &  97.5  & 95.9  & 96.0 & 96.1 &      87.8  &  97.2  &   97.5  &  97.7  \\
     CKD              &         300 &        100 &           2 &        12 &         9 &  99.2  & 96.7  & 97.8 & 96.6 &     100    &  99.4  &   99.5  &  99.1  \\
     Cirrhosis        &         203 &         68 &           3 &        10 &         5 &  46.8  & 50.7  & 51.5 & 45.7 &      46.9  &  50.9  &   47.5  &  54.0    \\
     Compas           &        3958 &       1320 &           2 &         5 &         8 &  65.5  & 63.7  & 64.9 & 62.5 &      68.1  &  66.6  &   67.1  &  67.6  \\
     CreditFraud      &        1107 &        369 &           2 &        30 &         0 &  93.3  & 92.0  & 93.1 & 91.0 &      94.0  &  93.0    &   94.1  &  94.3  \\
     Diabetes         &         292 &         98 &           2 &        13 &         1 &  71.3  & 85.6  & 86.1 & 80.8 &      80.2  &  78.7  &   83.0    &  88.0    \\
     EyeMovements     &        8202 &       2734 &           3 &        20 &         3 &  58.1  & 44.3  & 45.4 & 45.2 &      53.9  &  52.1  &   54.2  &  55.0    \\
     HAR              &        7352 &       2947 &           6 &        63 &         0 &  78.1  & 70.3  & 75.6 & 81.4 &      72.4  &  90.8  &   81.8  &  87.1  \\
     Heart            &         219 &         74 &           2 &         5 &        14 &  80.1  & 71.8  & 81.0 & 83.2 &      80.5  &  80.2  &   74.6  &  81.5  \\
     HeartFailure     &         224 &         75 &           2 &         6 &         5 &  50.9  & 65.2  & 68.6 & 65.3 &      64.4  &  59.7  &   62.6  &  68.4  \\
     HepatitisC       &         461 &        154 &           2 &        11 &         1 &  86.6  & 90.7  & 93.2 & 88.8 &      93.9  &  91.3  &   86.8  &  91.1  \\
     Letter           &       15000 &       5000 &          26 &        16 &         0 &  94.9  & 39.7  & 33.6 & 64.1 &      41.2  &  94.7  &   42.2  &  69.1  \\
     Liver            &         434 &        145 &           2 &         9 &         1 &  57.4  & 63.9  & 66.2 & 69.2 &      55.5  &  69.1  &   66.5  &  66.5  \\
     Mushroom         &        5898 &       1966 &           2 &         0 &        67 & 100    & 97.6  & 98.6 & 97.9 &     100    & 100    &  100    & 100    \\
     PDAC            &         516 &        172 &           2 &        31 &         0  &  49.5  & 52.6  & 60.9 & 62.1 &      57.7  &  58.4  &   54.4  &  59.6  \\
     SatIm            &        4435 &       2000 &           6 &        36 &         0 &  89.1  & 79.5  & 81.9 & 78.9 &      71.9  &  89.0    &   84.9  &  86.4  \\
     Stroke           &        3831 &       1278 &           2 &         3 &        11 &  52.4  & 73.4  & 75.9 & 69.8 &      50.5  &  62.3  &   73.4  &  76.2  \\
     TelcoChurn       &        5262 &       1754 &           2 &         2 &        21 &  70.2  & 73.3  & 75.0 & 74.6 &      70.9  &  68.7  &   74.5  &  75.9  \\
     \hline
     Average Rank     &          &         &          &        &        &   5.2 &  6.1 &  4.0   &  5.3 &       5.1 &   4.2 & 3.9 &   2.3 \\
    \hline
    \end{tabular}
}
\end{table*}

\subsection{Concatenating Inputs} \label{method-concat}
The Wide \& Deep Model~\cite{cheng2016wide} work demonstrates the effectiveness of bringing some inputs 
closer to the output layer. We adopt this method by concatenating the ThresholdLayer to every intermediate LogicLayer, 
i.e., from the second layer to the second last. The concatenated ThresholdLayer may either be the same layer 
reused at multiple places or different ThresholdLayers in each place. Experimental results show that using 
different input layers is better than using the same layer, and both approaches are better than not 
concatenating. We set multiple input concatenations as the default and present the results of other mentioned 
methods in Sec.~\ref{exp-ablation}.

\subsection{Model Simplification}
After training, we obtain logic expressions from the network and simplify them using SimPy~\cite{10.7717/peerj-cs.103}. 
We have found that this step is generally beneficial for reducing model size because our trainable 
ThresholdLayer acts as a feature selector. We also explore other simplification methods such as pruning 
and present the results in the ablation studies section, Sec.~\ref{exp-ablation}. Based on the results, 
we find that SimPy is sufficient for model simplification.

\section{Experiments}\label{sec-experiments}
We test our method and compare it with seven other methods on 20 tabular datasets. The seven methods 
we test are k-nearest neighbors (KNN), decision tree (DT), random forest (RF), Gaussian Naive Bayes (NB), 
AdaBoost (AB), MLP, and tree-initialized DDLGN (abbreviated to LGN in our tables). We list the characteristics 
of the datasets after preprocessing in Table~\ref{datasets-stats}. Their sample sizes range from hundreds to 
thousands, and their number of classes ranges from two to 26, which is typical for tabular datasets.

\subsection{Experimental Setup}
We first preprocess the datasets. This involves removing samples with too many missing values, one-hot encoding 
categorical features, and scaling continuous features using Min-Max scalers to ensure they range between \(0\) and \(1\). 
For each (dataset, model) pair, 
we conduct four experiments, each with a different random seed. Each experiment consists of three steps: 
hyperparameter search, training, and evaluation. We sample \(32\) parameter sets for every hyperparameter 
search and select the set with the best cross-validated balanced-class test accuracy. The number of folds 
ranges from two to four, with datasets containing fewer samples having more folds. We employ the Optuna 
search algorithm~\cite{optuna_2019} for neural network-based models (i.e., MLP, DDLGN, DLN) and use random 
sampling for traditional methods (i.e., KNN, DT, RF, NB, AB). We use the scikit-learn package~\cite{scikit-learn} 
for data processing and experiments with traditional methods. We train NN models using PyTorch~\cite{NEURIPS2019_9015}. 
We use Ray~\cite{moritz2018ray} to run parallel hyperparameter searches.

\begin{table*}[!ht]
    \centering
    \caption{Average number of operations for inference, assuming float16 for floating-point and int16 for integer.} \label{results-ops}
    \footnotesize{
    \begin{tabular}{c|cccccccc|cccccccc}
     & \multicolumn{8}{c|}{High-level OPs} & \multicolumn{8}{c}{Basic logic gate-level OPs} \\
     & KNN & DT & RF & NB & AB & MLP & LGN & DLN    & KNN & DT & RF & NB & AB & MLP & LGN & DLN\\
    \hline
     Bank        & 316K  &    2 &  515 & 72   &      300   & 1.1K  & 251     &  52    & 106M    & 366    & 68.3K  & 323K   & 143K         & 696K    & 3.8K      & 1.9K    \\
     Breast     & 21.5K &    3 &  578 & 210  &      378   & 6.7K  & 126     &  61    & 8.2M    & 503    & 81.2K  & 386K   & 180K         & 4.1M    & 545       & 256     \\
     Cardio & 162K  &    5 &  836 & 378  &      478   & 10.3K & 721     & 211    & 37.4M   & 915    & 129K   & 608K   & 227K         & 6.3M    & 10.6K     & 6.1K    \\
     CKD              & 21.1K &    3 &  614 & 132  &      426   & 3.2K  & 351     &  91    & 9.0M    & 412    & 86.2K  & 350K   & 203K         & 1.9M    & 7.4K      & 4.4K    \\
     Cirrhosis        & 9.0K  &    3 &  157 & 144  &      389   & 1.7K  & 324     & 120    & 2.7M    & 412    & 7.8K   & 501K   & 185K         & 1.0M    & 5.6K      & 6.0K    \\
     Compas           & 185K  &    2 &  404 & 84   &      419   & 1.1K  & 179     &  60    & 64.1M   & 229    & 48.0K    & 328K   & 200K         & 647K    & 2.8K      & 2.7K    \\
     Credit      & 76.5K &    2 &  636 & 186  &      437   & 5.9K  & 755     & 169    & 10.1M   & 366    & 87.5K  & 375K   & 208K
& 3.6M    & 17.0K       & 10.5K   \\
     Diabetes         & 14.4K &    1 &  150 & 90   &      328   & 1.1K  & 369     &  44    & 6.0M    & 183    & 10.5K  & 331K   & 156K         & 667K    & 5.8K      & 3.2K    \\
     EyeMove     & 476K  &    2 &  417 & 216  &      478   & 9.6K  & 545     & 297    & 65.3M   & 366    & 51.8K  & 534K   & 227K         & 5.9M    & 10.7K     & 10.2K   \\
     HAR              & 1.4M  &    4 &  730 & 1.1K &      542   & 40.5K & 1.9K    & 392    & 520M    & 686    & 108K   & 1.4M   & 258K         & 25.1M   & 29.2K     & 31.4K   \\
     Heart            & 10.8K &    3 &  394 & 120  &      457   & 1.9K  & 167     &  80    & 2.5M    & 412    & 48.3K  & 345K   & 218K         & 1.2M    & 2.8K      & 1.3K    \\
     HeartFail     & 7.7K  &    3 &  189 & 72   &      376   & 754   & 149     &  48    & 2.3M    & 458    & 13.7K  & 323K   & 179K         & 458K    & 2.9K      & 2.4K    \\
     Hepatitis       & 14.7K &    3 &  486 & 78   &      527   & 1.7K  & 190     &  57    & 2.1M    & 549    & 66.7K  & 326K   & 251K         & 1.1M    & 6.2K      & 4.1K    \\
     Letter           & 853K  &    6 &  524 & 1.3K &      523   & 15.6K & 599     & 189    & 297M    & 1.0K   & 71.4K  & 4.4M   & 249K         & 9.6M    & 8.7K      & 22.4K   \\
     Liver            & 14.2K &    2 &  179 & 66   &      438   & 606   & 192     &  39    & 4.2M    & 366    & 13.5K  & 320K   & 209K         & 366K    & 4.6K      & 3.2K    \\
     Mush.         & 1.1M  &    3 &  779 & 408  &      509   & 25.5K & 350     & 320    & 332M    & 458    & 117K   & 476K   & 242K         & 15.8M   & 1.2K      & 1.5K    \\
     PDAC            & 52.1K &    1 &  130 & 192  &      368   & 7.4K  & 496     &  59    & 22.7M   & 137    & 3.0K     & 378K   & 175K         & 4.6M    & 18.9K     & 5.0K    \\
     SatIm            & 448K  &    6 &  880 & 666  &      435   & 14.6K & 858     & 422    & 138M    & 1.0K   & 135K   & 1.2M   & 207K         & 9.0M    & 16.2K     & 25.9K   \\
     Stroke           & 176K  &    2 &  612 & 90   &      522   & 1.9K  & 194     &  40    & 51.2M   & 229    & 86.1K  & 331K   & 249K         & 1.1M    & 1.9K      & 1.6K    \\
     Telco       & 362K  &    3 &  551 & 144  &      384   & 3.3K  & 182     &  67    & 109M    & 503    & 74.8K  & 356K   & 183K         & 2.0M    & 1.6K      & 1.3K    \\
     \hline
     Rank     & 8.0     &    1.0 &    5.0 & 3.4  &        4.9 & 7.0     & 4.7    &   2.1 & 8.0       & 1.1   & 3.9    & 6.0      &
5.0            & 7.0       & 2.8       & 2.3    \\
    \hline
    \end{tabular}
}
\end{table*}

\begin{table*}[!ht]
    \centering
    \caption{Average number of model parameters and disk space in Bytes, assuming float16 for numerical values and 
int16 for indices.} \label{results-params}
    \footnotesize{
    \begin{tabular}{c|cccccccc|cccccccc}
     & \multicolumn{8}{c|}{Number of parameters} & \multicolumn{8}{c}{Disk space in Bytes} \\
     & KNN & DT & RF & NB & AB & MLP & LGN & DLN    & KNN & DT & RF & NB & AB & MLP & LGN & DLN\\
    \hline
     Bank        & 90.0K   &   13 & 3.0K &   46 & 702        & 598   & 586     & 129   & 180K    &     25 & 5.9K   & 92     & 1.4K         & 1.2K    & 1.2K      & 258     \\
     Breast     & 7.5K  &   16 & 3.6K &  138 & 886        & 3.4K  & 271     & 156   & 15.0K     &     32 & 7.2K   & 276    & 1.8K         & 6.8K    & 542       & 313     \\
     Cardio & 66.9K &   29 & 5.2K &  249 & 1.1K       & 5.2K  & 1.7K    & 582   & 134K    &     57 & 10.3K  & 498    & 2.2K         & 10.4K   & 3.4K      & 1.2K    \\
     CKD              & 6.6K  &   14 & 3.1K &   86 & 998        & 1.6K  & 834     & 232   & 13.2K   &     27 & 6.1K   & 172    & 2.0K           & 3.2K    & 1.7K      & 465     \\
     Cirrhosis        & 3.2K  &   24 & 290  &   93 & 910        & 880   & 767     & 298   & 6.5K    &     47 & 580    & 186    & 1.8K         & 1.8K    & 1.5K      & 596     \\
     Compas           & 55.4K &    8 & 1.8K &   54 & 980        & 554   & 414     & 136   & 111K    &     15 & 3.6K   & 108    & 2.0K         & 1.1K    & 828       & 271     \\
     Credit      & 34.3K &   14 & 3.8K &  122 & 1.0K       & 3.0K  & 1.8K    & 413   & 68.6K   &     27 & 7.5K   & 244    & 2.0K         & 6.0K    & 3.7K      & 826     \\
     Diabetes         & 4.4K  &    6 & 354  &   58 & 766        & 568   & 872     & 99    & 8.8K    &     12 & 708    & 116    & 1.5K         & 1.1K    & 1.7K      & 196     \\
     EyeMove     & 197K  &   11 & 1.9K &  141 & 1.1K       & 4.8K  & 1.3K    & 989   & 394K    &     22 & 3.7K   & 282    & 2.2K         & 9.7K    & 2.6K      & 2.0K    \\
     HAR              & 471K  &   39 & 6.2K &  762 & 1.3K       & 20.4K & 4.5K    & 1.1K  & 941K    &     77 & 12.5K  & 1.5K   & 2.5K         & 40.8K   & 9.1K      & 2.3K    \\
     Heart            & 4.4K  &   19 & 1.6K &   78 & 1.1K       & 978   & 391     & 209   & 8.8K    &     37 & 3.2K   & 156    & 2.1K         & 2.0K    & 782       & 418     \\
     HeartFail     & 2.7K  &   14 & 451  &   46 & 878        & 392   & 354     & 111   & 5.4K    &     27 & 902    & 92     & 1.8K         & 784     & 708       & 222     \\
     Hepatitis       & 6.0K  &   21 & 2.7K &   50 & 1.2K       & 900   & 475     & 127   & 12.0K     &     42 & 5.4K   & 100    & 2.5K         & 1.8K    & 950       & 254     \\
     Letter           & 255K  &  101 & 6.1K &  858 & 1.2K       & 7.9K  & 1.4K    & 1.3K  & 510K    &    202 & 12.2K  & 1.7K   & 2.5K         & 15.9K   & 2.7K      & 2.6K    \\
     Liver            & 4.8K  &   15 & 440  &   42 & 1.0K       & 318   & 468     & 98    & 9.6K    &     30 & 880    & 84     & 2.0K         & 635     & 936       & 196     \\
     Mush.         & 401K  &   18 & 4.9K &  270 & 1.2K       & 12.9K & 766     & 1.3K  & 802K    &     35 & 9.8K   & 540    & 2.4K         & 25.7K   & 1.5K      & 2.5K    \\
     PDAC            & 16.5K &    5 & 158  &  126 & 861        & 3.7K  & 1.3K    & 131   & 33.0K     &     10 & 316    & 252    & 1.7K         & 7.5K    & 2.5K      & 262     \\
     SatIm            & 164K  &   48 & 6.4K &  438 & 1.0K       & 7.4K  & 2.0K    & 1.2K  & 328K    &     95 & 12.8K  & 876    & 2.0K         & 14.8K   & 4.1K      & 2.5K    \\
     Stroke           & 57.5K &    8 & 3.6K &   58 & 1.2K       & 964   & 444     & 99    & 115K    &     15 & 7.3K   & 116    & 2.4K         & 1.9K    & 888       & 197     \\
     Telco       & 126K  &   18 & 3.0K &   94 & 900        & 1.7K  & 410     & 167   & 253K    &     35 & 6.1K   & 188    & 1.8K         & 3.4K    & 820       & 334     \\
     \hline
     Rank     & 8.0     &    1.0 & 6.1 &    2.0 & 5.2       & 5.9   & 4.7    & 3.3  & 8.0       &      1.0 & 6.1   & 2.0      &
5.2         & 5.9     & 4.7      & 3.3    \\
    \hline
    \end{tabular}
}
\end{table*}

\subsection{Accuracy}
We measure the predictive power of the models using balanced-class accuracy. Table~\ref{results-acc} presents 
the average results. DLN achieved the best average rank of \(2.3\) and NN-based methods generally 
outperform traditional methods. Another observation is that bagging and boosting help, as random forest 
and AdaBoost outperform the decision tree, as expected. DDLGN requires binary inputs; hence, continuous features need to be 
binned before passing to the network. In the original work~\cite{petersen2022deep}, DDLGN bins are selected 
manually for tabular data and uniformly for image data. We found that, compared to uniform thresholds, 
tree-based binning~\cite{gorishniy2022embeddings} improves DDLGN accuracy by a significant margin. Thus, we present results 
for tree-initialized DDLGNs.

\subsection{Efficiency}
We measure efficiency from two perspectives: inference cost and disk space cost. For inference computation cost,
we provide the number of high-level operations (OPs) and the corresponding number of basic logic gate-level 
operations in Table~\ref{results-ops}. High-level operations include addition, comparison, multiplication, 
division, logarithm, etc. For each high-level operation, depending on the input type (e.g., float32, int64),
we can compute its total number of operations in basic hardware logic gates. For example, to perform addition 
of two \(n\)-bit inputs, the least significant bit (LSB) is fed to a half-adder and the other bits to
full-adders. A half-adder can be implemented with \(5\) NAND gates and a full-adder with \(9\) NAND gates. Hence, adding 
two \(n\)-bit inputs requires \(5 + 9*(n-1)\) NAND operations. We can similarly convert all high-level operations to basic 
\(2\)-input logic operations. We count AND/OR/NAND/NOR as one logic operation, XOR/XNOR 
as three, and NOT as 0. We used the default number types when training models, i.e., float32 
for NNs and float64 for most traditional models. Since most models can maintain inference accuracy even 
after downgrading the datatype to float16, we assume that floating-point numbers have a data type of float16 
and integers have a data type of int16 when presenting the counts in Table~\ref{results-ops}. DLN would have a higher advantage
with larger datatypes because the only place it needs floating-point operations is when comparing continuous features 
with threshold values in the ThresholdLayer. Thus, this advantage of DLN is muted in Table~\ref{results-ops}. The table shows 
that decision trees are the most computationally efficient method and DLN is the second best. 
Logic-based networks are orders of magnitude more efficient than multiplication-based networks.

Table~\ref{results-params} presents the number of parameters and the required disk space for storage. Trees need 
to store threshold values and indices of features and child nodes. DLNs need to store the functionality of neurons 
and indices of their incoming links. We assumed that each float/int requires \(16\) bits and each index also needs 
\(16\) bits. On an average, decision trees are the smallest in size and naive Bayes the second smallest. The average rank of DLNs 
is \(3.3\), which is the smallest among NNs and substantially smaller than vanilla MLPs.

\subsection{Interpretability}
\begin{figure*}[!ht]
\centering
\includegraphics[width=\textwidth]{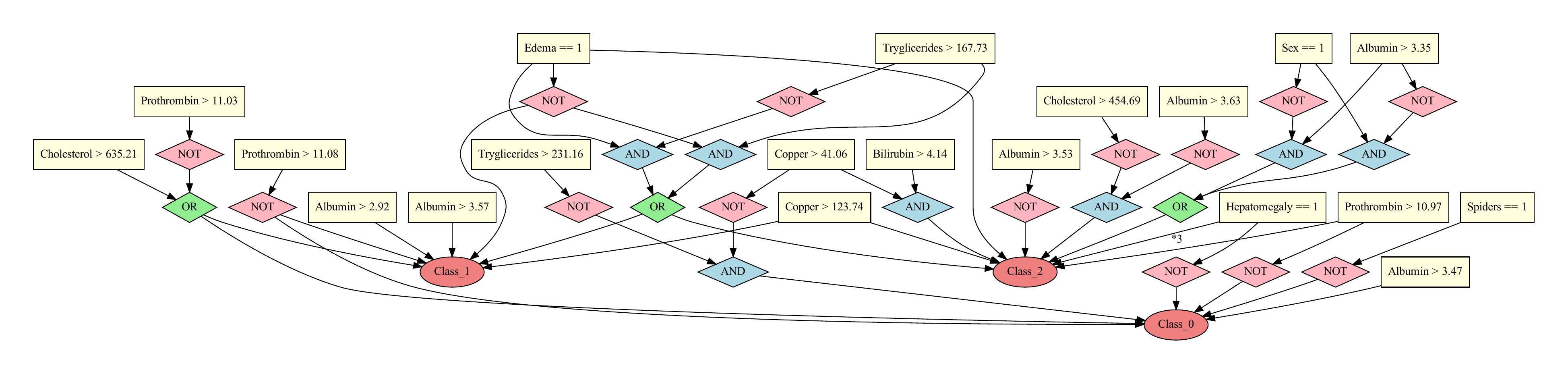}
\caption{Visualization of the decision-making process of a DLN for predicting cirrhosis disease. 
It achieves a balanced-class test accuracy of \(59.3\%\). The DLN uses \(6\) continuous features 
and \(4\) categorical features, selected from an original set of \(10\) continuous features 
and \(5\) categorical features.}
\label{viz-cirrhosis}
\end{figure*}

\begin{figure*}[!ht]
\centering
\includegraphics[width=\textwidth]{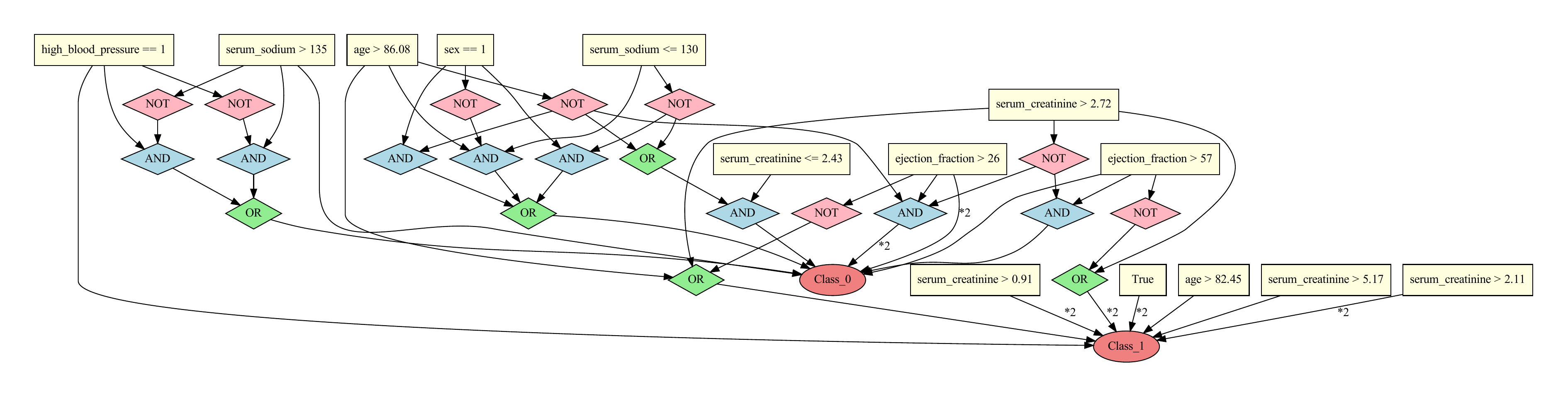}
\caption{Visualization of the decision-making process of a DLN for predicting heart failure disease. 
It achieves a balanced-class test accuracy of \(70.5\%\). The DLN uses \(4\) continuous features 
and \(2\) categorical features, selected from an original set of \(6\) continuous features 
and \(5\) categorical features.}
\label{viz-heartFailure}
\end{figure*}

\begin{figure*}[!ht]
\centering
\includegraphics[width=\textwidth]{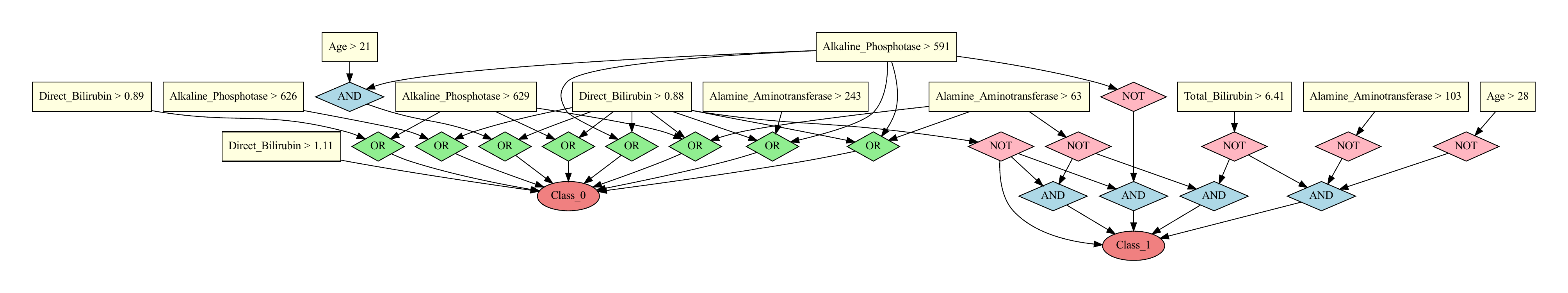}
\caption{Visualization of the decision-making process of a DLN for predicting liver disease. 
It achieves a balanced-class test accuracy of \(72.5\%\). The DLN uses \(5\) continuous features, 
selected from an original set of \(9\) continuous features and \(1\) categorical feature.}
\label{viz-liver}
\end{figure*}

DLNs are interpretable by nature and their interpretability is reinforced by the feature selection power of 
ThresholdLayers. Figs.~\ref{viz-cirrhosis}, \ref{viz-heartFailure}, and \ref{viz-liver} illustrate the 
decision-making process of DLNs. Take the DLN in Fig.~\ref{viz-cirrhosis} as an example. It achieves a balanced-class 
test accuracy of \(59.3\%\) on the Cirrhosis dataset and makes decisions based on \(6\) continuous features 
and \(4\) categorical features, chosen from \(10\) continuous and \(5\) categorical features. $\text{Class}_0$, 
$\text{Class}_1$, and $\text{Class}_2$ represent the three class types. Yellow boxes represent features and diamonds 
represent logic operators. The DLN binarizes inputs based on comparisons for continuous features and equality checks 
for categorical features. Logic operators receive binary inputs and output binary values. Class nodes receive 
binary scores from either feature boxes or logic operators and aggregate the scores. A link to a class 
node represents a logic rule. The multiplier \(*\) on the right of the link to the class node represents 
the strength of the associated rule. The class with the highest score is the final prediction.

\begin{table*}[!ht]
    \centering
    \caption{Average balanced-class test accuracy percentage of different methods and their ranks based on the 
number of basic logic-gate operations.} \label{results-ablation}
    \footnotesize{
    % \begin{tabular}{c|cccccccccccc}
    \begin{tabular}{c*{12}{>{\centering\arraybackslash}m{0.75cm}}}
     & \multicolumn{1}{c|}{\multirow{2}*{Orig}} & \multicolumn{3}{c|}{Subspace} & \multicolumn{1}{c|}{STE} & \multicolumn{2}{c|}{Concat} & \multicolumn{1}{c|}{Phase} & \multicolumn{1}{c|}{Fixed} & \multicolumn{1}{c|}{\multirow{2}*{Gumbel}} & \multicolumn{1}{c|}{\multirow{2}*{Freeze}} & \multirow{2}*{Prune} \\
     & \multicolumn{1}{c|}{} & w/o & 16 & \multicolumn{1}{c|}{4} & \multicolumn{1}{c|}{w/o} & w/o & \multicolumn{1}{c|}{same} & \multicolumn{1}{c|}{w/o} & \multicolumn{1}{c|}{Thresh} & \multicolumn{1}{c|}{} & \multicolumn{1}{c|}{} & \\
    \hline
     BankChurn        &  75.8  &    75.7  &     76.4 &    76.4 &    76.9  &       75.3  &         74.1  &      77.1  &      76.5  &    76.1  &    75.2  &    76.7 \\
     BreastCancer     &  65.1  &    63.8  &     63.6 &    64.1 &    59.5  &       63.9  &         65.0    &      63.8  &      64.7  &    62.6  &    66.8  &    65.0   \\
     Cardiotocography &  97.7  &    97.3  &     97.6 &    97.1 &    97.7  &       97.8  &         97.8  &      97.7  &      97.7  &    97.4  &    97.6  &    97.7 \\
     CKD              &  99.1  &    99.6  &     99.6 &    98.3 &    99.2  &       99.1  &         99.1  &      99.6  &      98.5  &    99.6  &    99.8  &    99.0   \\
     Cirrhosis        &  54.0    &    51.7  &     51.3 &    52.6 &    51.6  &       49.7  &         54.2  &      53.8  &      49.6  &    56.0    &    53.4  &    54.9 \\
     Compas           &  67.6  &    67.5  &     67.6 &    67.2 &    67.2  &       67.2  &         67.2  &      66.6  &      68.0    &    67.1  &    67.5  &    67.4 \\
     CreditFraud      &  94.3  &    93.7  &     93.6 &    93.9 &    94.1  &       94.5  &         94.5  &      93.1  &      93.0    &    94.1  &    95.0    &    94.1 \\
     Diabetes         &  88.0    &    84.5  &     81.1 &    85.5 &    84.7  &       85.1  &         83.5  &      83.2  &      86.8  &    85.6  &    85.0    &    80.5 \\
     EyeMovements     &  55.0    &    55.2  &     55.5 &    55.2 &    51.2  &       54.5  &         54.0    &      53.3  &      56.9  &    55.0    &    54.8  &    55.3 \\
     HAR              &  87.1  &    85.7  &     87.2 &    87.6 &    86.7  &       84.0    &         85.4  &      87.2  &      87.5  &    87.1  &    86.8  &    86.8 \\
     Heart            &  81.5  &    80.0    &     79.7 &    81.2 &    82.3  &       79.0    &         81.9  &      82.0    &      78.3  &    77.5  &    83.1  &    80.2 \\
     HeartFailure     &  68.4  &    69.8  &     69.9 &    71.9 &    68.5  &       71.7  &         65.8  &      71.0    &      66.8  &    76.0    &    67.4  &    67.3 \\
     HepatitisC       &  91.1  &    92.1  &     90.9 &    91.5 &    89.0    &       93.4  &         92.4  &      91.2  &      88.5  &    94.4  &    93.3  &    91.7 \\
     Letter           &  69.1  &    68.2  &     66.6 &    68.7 &    67.2  &       63.5  &         65.3  &      63.9  &      71.5  &    67.6  &    69.8  &    68.1 \\
     Liver            &  66.5  &    68.1  &     67.7 &    68.2 &    68.5  &       67.5  &         68.0    &      66.5  &      65.9  &    67.9  &    68.0    &    66.3 \\
     Mushroom         & 100    &   100    &    100   &    99.9 &   100    &       99.9  &        100    &     100    &     100    &   100    &   100    &    99.8 \\
     PDAC            &  59.6  &    59.5  &     56.3 &    57.1 &    58.1  &       59.0    &         59.9  &      58.6  &      54.9  &    58.7  &    57.7  &    58.7 \\
     SatIm            &  86.4  &    85.7  &     85.7 &    86.2 &    82.5  &       84.6  &         85.1  &      86.9  &      86.5  &    86.0    &    86.4  &    85.5 \\
     Stroke           &  76.2  &    75.4  &     75.3 &    74.3 &    73.9  &       75.6  &         75.6  &      74.7  &      74.5  &    73.1  &    75.0    &    74.7 \\
     TelcoChurn       &  75.9  &    75.2  &     75.7 &    74.9 &    75.1  &       75.5  &         74.7  &      75.0    &      75.9  &    75.1  &    75.4  &    75.7 \\
     \hline
     Accuracy Rank         &   2.3 &     2.6 &      2.7 &     2.7 &     2.9 &        2.8 &          2.6 &
2.6 &       3.2 &     2.5 &     2.3 &     2.7 \\
     \hline
     \hline
     OPs Rank          &   2.3 &     2.1 &      2.2 &     2.5 &     2.2  &        2.0    &          2.1
&       2.5 &       2.9  &     2.2  &     2.3 &     2.1 \\
    \hline
    \end{tabular}
}
\end{table*}

\subsection{Ablation Studies} \label{exp-ablation}
Next, we present ablation studies based on subsetting of the search space, use of STE, concatenation of input 
and LogicLayers, simultaneous optimization of neurons and links, application of fixed thresholds, use of 
Gumbel-Softmax, progressive freezing, and pruning. In Table~\ref{results-ablation}, we show the average DLN 
accuracy and rankings based on both accuracy and the number of basic logic gate-level operations. 
Rank is derived based on placing the corresponding column in Tables~\ref{results-acc} 
and~\ref{results-ops}. The DLN results are reproduced in the first column from those two tables. 
These original DLN results are derived under subsetting both gate and link 
search space to \(8\), using STE in all layers, and concatenating different ThresholdLayers to all middle 
LogicLayers. In the other columns, we list differences relative to the original 
setting. For freezing and pruning, we run a \(32\)-sample hyperparameter search based on the 
hyperparameters of the original model. We search over only the parameters related to freezing and 
pruning: starting epoch for freezing, starting and ending epochs, and convergence threshold for 
pruning. For other cases, we perform hyperparameter search, training, and evaluation 
similar to the one performed for the original model. \\
\textbf{Subset Search Space:} We present results of searching over the full space, using a link 
subspace of size \(16\) for inputs of each neuron, and using both a link and gate subspace of size \(4\). We 
found that size \(8\) represents a local optimum. \\
\textbf{STE:} We show the results of when STE is not applied. On average, these results are worse. \\
\textbf{Input Concatenation:} We test not concatenating the ThresholdLayer with the LogicLayer and concatenating 
the same ThresholdLayer to every middle LogicLayer. We conclude that using different ThresholdLayers is better 
than using the same ThresholdLayer, and that itself is better than not concatenating. \\
\textbf{Unified Phase:} The results indicate that optimizing neuron functionalities and link 
connections in an alternating fashion is better in terms of both accuracy and model size than optimizing them 
together. \\
\textbf{Fixed ThresholdLayer:} We present results when the ThresholdLayers are initialized and frozen, 
i.e., only LogicLayers and SumLayers are trained. This not only reduces model accuracy but also 
results in larger models. Thus, it is better to train the ThresholdLayers. \\
\textbf{Gumbel-Softmax:} We replace Softmax activations with Gumbel-Softmaxes, using a 
Gumbel noise scale of \(0.5\). However, this does not improve the model. \\
\textbf{Progressive Freezing:} Based on the observation that layers closer to inputs perform more 
pattern extraction and layers closer to outputs perform more abstract pattern aggregation, we experiment with
progressively freezing layer parameters to enable better gradient flow as training progresses. Results 
show that progressive freezing does not significantly improve model performance. \\
\textbf{Progressive Pruning:} Some neurons are not helpful in making predictions, e.g., a neuron 
ANDed with an always False neuron. Neurons that do not converge well may be redundant or even 
harmful to prediction accuracy. Hence, we test pruning out neurons that do not converge well (i.e., probabilities 
do not concentrate on one type of gate) in a progressive manner. However, this does not improve model 
accuracy. One reason may be that a larger DLN architecture search space is needed when applying 
pruning. Another reason may be that a non-useful neuron is already ignored in the link search phase.

\subsection{Limitations}
A shortcoming of a DLN is its training cost. To train a logic network, gradient descent must traverse
layers of Softmax functions, which results in both slower backpropagation and vanishing gradients. Techniques 
like subsetting the search space and breaking the optimization problem into two phases help with 
training but do not solve the problem completely. Another limitation is that a DLN does not always perform 
well. For example, in Table~\ref{results-acc}, on the Letter dataset, DLN underperforms MLP and some traditional 
models by a large margin. This may be due to the fact that rule-based approaches are not suitable for the Letter 
dataset since tree-based models also do not perform well.

\section{Conclusion and Future Directions} \label{sec-conclusion}
We introduced a methodology for training DLNs whose inference processes
can be explained through logic rules. The training methodology involves two iterative phases: 
one for determining neuron functions and another for determining neuron connections.
We demonstrated that DLNs learned using this methodology are accurate, interpretable, and efficient.
In the future, we plan to explore DLNs in the context of ensemble frameworks such as bagging and boosting.
This would involve adding aggregation sources for class nodes.  We will also explore incorporation of prior 
rule-based knowledge into DLNs through addition of logic paths from class nodes to the rules
or representing these rules using trainable neurons.

\bibliographystyle{IEEEtran}
\bibliography{IEEEabrv, references}

\end{document}